\documentclass[sigconf,natbib=false]{acmart}
\usepackage{multirow}
\usepackage{booktabs}
\usepackage{colortbl} % 引入colortbl包以便对表格颜色进行操作

\AtBeginDocument{%
  }

\copyrightyear{2024}
\acmYear{2024}
\setcopyright{acmlicensed}
\acmConference[MM '24]{Proceedings of the 32nd ACM International Conference on Multimedia}{October 28-November 1, 2024}{Melbourne, VIC, Australia}
\acmBooktitle{Proceedings of the 32nd ACM International Conference on Multimedia (MM '24), October 28-November 1, 2024, Melbourne, VIC, Australia}
\acmDOI{xxx}
\acmISBN{xxx}

\renewcommand\footnotetextcopyrightpermission[1]{} % removes footnote with conference information in first column
\RequirePackage[
  datamodel=acmdatamodel,
  style=acmnumeric,
  ]{biblatex}
\begin{document}

\title[A Unified Understanding of Adversarial Vulnerability]{A Unified Understanding of Adversarial Vulnerability Regarding Unimodal Models and Vision-Language Pre-training Models}

\author{Haonan Zheng}
\email{zhenghaonan@mail.nwpu.edu.cn}
\orcid{0009-0009-2017-962X}
\affiliation{%
  \institution{Northwestern Polytechnical University}
  \department{School of Electronics and Information}
  \city{Xi'an}
  \state{Shaanxi}
  \country{China}
}

\author{Xinyang Deng}
% \email{xinyang.deng@nwpu.edu.cn}
% \orcid{0000-0001-8181-7001}
\affiliation{%
  \institution{Northwestern Polytechnical University}
  \department{School of Electronics and Information}
  \city{Xi'an}
  \state{Shaanxi}
  \country{China}
}

\author{Wen Jiang}
% \authornotemark[2]
\authornote{Corresponding author}
% \email{jiangwen@nwpu.edu.cn}
% \orcid{0000-0001-5429-2748}
\affiliation{%
  \institution{Northwestern Polytechnical University}
  \department{School of Electronics and Information}
  \city{Xi'an}
  \state{Shaanxi}
  \country{China}
}

\author{Wenrui Li}
\email{liwr618@163.com}
% \orcid{0000-0002-2393-9016}
\affiliation{%
  \institution{Harbin Institute of Technology}
  \department{Department of Computer Science and Technology}
  \city{Harbin}
  % \state{Shaanxi}
  \country{China}
}

\renewcommand{\shortauthors}{Haonan Zheng, Xinyang Deng, Wen Jiang, and Wenrui Li}

\begin{abstract}

With Vision-Language Pre-training (VLP) models demonstrating powerful multimodal interaction capabilities, the application scenarios of neural networks are no longer confined to unimodal domains but have expanded to more complex multimodal V+L downstream tasks.
% √
The security vulnerabilities of unimodal models have been extensively examined, whereas those of VLP models remain challenging.
% √
We note that in CV models, the understanding of images comes from annotated information, while VLP models are designed to learn image representations directly from raw text.
% √
Motivated by this discrepancy, we developed the Feature Guidance Attack (FGA), a novel method that uses text representations to direct the perturbation of clean images, resulting in the generation of adversarial images.
% √
FGA is orthogonal to many advanced attack strategies in the unimodal domain, facilitating the direct application of rich research findings from the unimodal to the multimodal scenario.
% √
By appropriately introducing text attack into FGA, we construct Feature Guidance with Text Attack (FGA-T). 
% √
Through the interaction of attacking two modalities, FGA-T achieves superior attack effects against VLP models.
% √
Moreover, incorporating data augmentation and momentum mechanisms significantly improves the black-box transferability of FGA-T. 
% √
% We conduct experiments on four VLP models and six V+L downstream tasks, demonstrating that FGA-T surpasses current leading methods in white-box by a considerable margin and black-box scenarios .
Our method demonstrates stable and effective attack capabilities across various datasets, downstream tasks, and both black-box and white-box settings, offering a unified baseline for exploring the robustness of VLP models.
% \textbf{Our codes will be available after acceptance.}

\end{abstract}

%    <concept>
%        <concept_id>10002978</concept_id>
%        <concept_desc>Security and privacy</concept_desc>
%        <concept_significance>500</concept_significance>
%        </concept>
%    <concept>
%        <concept_id>10002951.10003317.10003371.10003386</concept_id>
%        <concept_desc>Information systems~Multimedia and multimodal retrieval</concept_desc>
%        <concept_significance>500</concept_significance>
%        </concept>
%  </ccs2012>

% \ccsdesc[500]{Security and privacy}
% \ccsdesc[500]{Information systems~Multimedia and multimodal retrieval}



\maketitle

\section{Introduction}

ViT provides an effective Transformer-based encoder for the visual modality \cite{vit}, ensuring the feature extraction of multimodal input through a unified encoding manner, significantly advancing the Vision-and-Language tasks \cite{vqa1, vln1, vln2, itr1}. 
Various VLP models \cite{albef,tcl,beit3,clip} continually improve performance in V+L downstream tasks through diverse pre-training tasks and architectural designs \cite{beit,beit3,vilt,li1,li2}. 
However, the previous research in unimodal fields such as Computer Vision (CV) and Natural Language Processing (NLP) highlights the vulnerability of neural networks to adversarial attacks \cite{fgsm,bertatk}. 
Although adversarial robustness, particularly in CV, has been extensively explored in terms of attack strategies \cite{cw, deepfool}, defence mechanisms \cite{pgd}, and transferability \cite{mi,vmi}, the study of adversarial robustness in VLP models remains challenges \cite{arra,coattack,advclip,sga}. 
Our study aims to develop a unified architecture to explore commonalities between multimodal and unimodal tasks from the perspective of adversarial attacks. 
In other words, we seek to bridge the gap, allowing rich findings in unimodal adversarial robustness to be directly applied to the multimodal scenario.

The first question we consider is ``Which modality should be paid more attention?'' 
We primarily focus on perturbations in the image modality, with perturbations in the text modality serving as orthogonal 
(1) \textbf{Semantic consistency}: Visual adversarial examples maintain semantic consistency, i.e., noise addition within reasonable limits doesn't change human comprehension. 
Conversely, text adversarial examples risk semantic distortion, potentially introducing spelling errors. 
(2) \textbf{Differentiability}: Image inputs are continuous and differentiable, unlike text tokens which are discrete and non-differentiable making text-only attacks less effective.
(3) \textbf{Accessibility}: In real-world scenarios, text often serves as the primary means of interaction between users and AI models, with limited opportunities for attackers to modify user-generated text. 
In contrast, models can automatically acquire image data, simplifying the process for attackers to introduce perturbations.
In fact, \cite{sga} also primarily focuses on enhancing attack strategies in the visual modality to improve adversarial transferability and \cite{advclip} does not involve text attacks.

\begin{figure}[t]
  \centering
  \includegraphics[width=\linewidth]{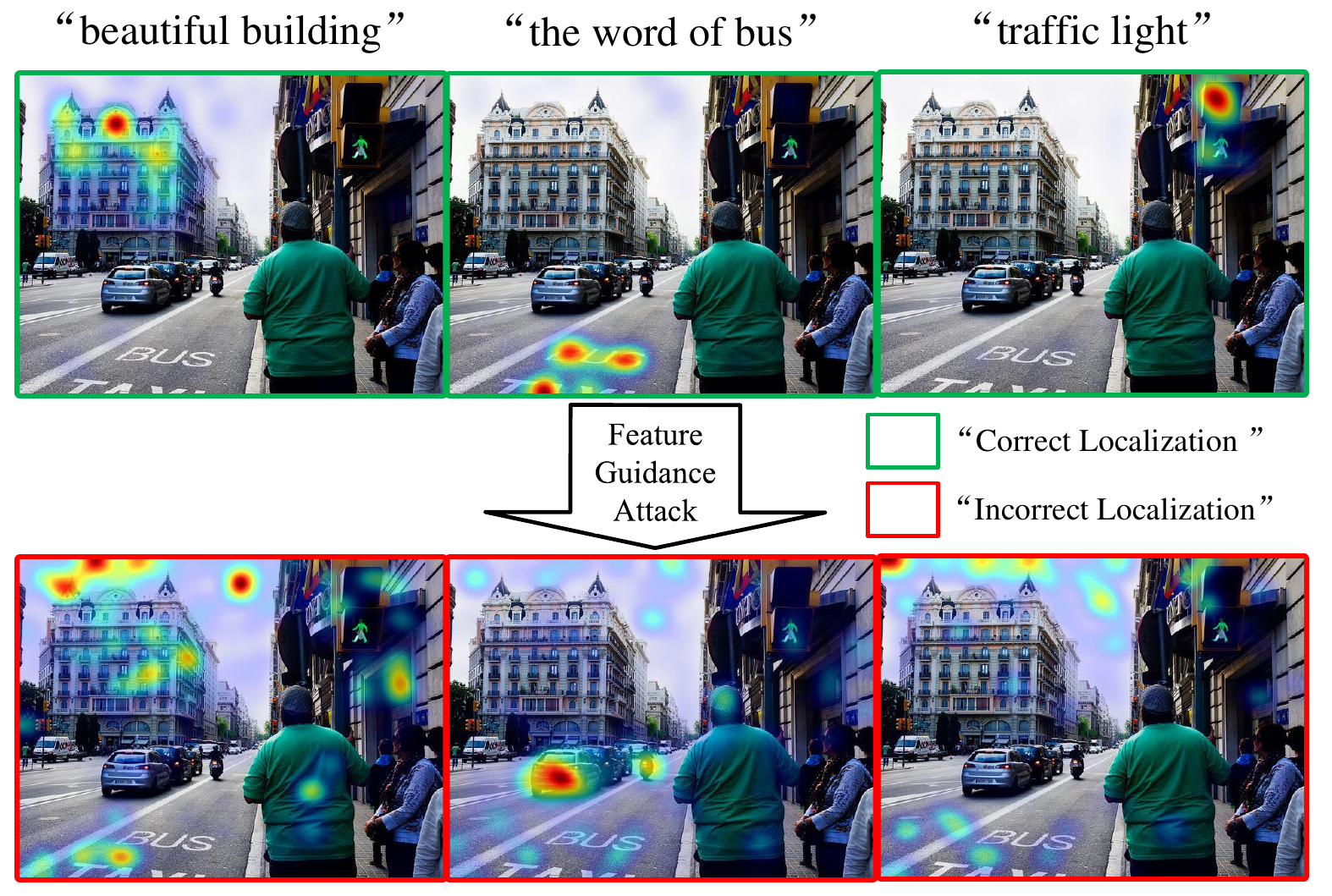}
  \caption{ALBEF computes Grad-CAM\cite{grad-cam} visualizations on the self-attention maps. Before FGA, ALBEF can accurately localize image content based on textual cues. After FGA, ALBEF's understanding of the image becomes confused.}
  \Description{ALBEF computes Grad-CAM\cite{grad-cam} visualizations on the self-attention maps. Before FGA, ALBEF can accurately localize image content based on textual cues. After FGA, ALBEF's understanding of the image becomes confused.}
\end{figure}

The second question is ``How to unify multimodal and unimodal scenarios in exploring adversarial robustness?'' 
We conceptualize image adversarial attacks as a feature-guided process. 
For unimodal models which primarily learn to understand images through detailed annotation information (such as category labels), attacking an image involves steering its embedding away from the feature vector linked to its correct annotation \cite{fgsm}. 
This deviation induces a biased comprehension of the image within the network. 
Alternatively, the image embedding can be guided closer to the feature vector associated with an incorrect annotation, thereby leading the network to make a predetermined error \cite{targetedatk}.
In the multimodal scenario, models are encouraged to understand images from raw text, providing a broader and more accessible source of supervision \cite{clip}. 
This also offers more flexible guiding information for the adversarial attack. 
By guiding the image embedding away from the correct text description, we induce the VLP model to develop an incorrect understanding of the image itself. 
Similarly, directing the embedding towards an incorrect text description intentionally misleads the model into adopting a specific erroneous interpretation. 
This strategy is termed Feature Guidance Attack (FGA).
Expanding upon FGA, we employ adversarial texts from text attacks as guiding information to generate adversarial images, thus obtaining a novel multimodal attack. 
This approach exacerbates the model's misinterpretation called Feature Guidance with Text Attack (FGA-T).
Furthermore, we introduce additional orthogonal mechanisms to enhance the adversarial transferability of FGA-T in the black-box scenario.
Code: https://github.com/LibertazZ/FGA

Our contributions can be summarized as follows:
\begin{itemize}
\item 
We provide FGA, using original text as the supervision source for the adversarial attack on VLP models, inducing the network to misinterpret adversarial images.
\item 
We introduce cross-modal interaction through adversarial text, forming a novel multimodal adversarial attack that enhances white-box attack strength, and improves black-box transferability through additional mechanisms.
\item 
Our approach is theoretically orthogonal to any unimodal attack enhancement mechanism. Empirical evidence based on multiple datasets and VLP models demonstrates the broad applicability of our method to various V+L multimodal tasks, providing a unified baseline for the exploration of multimodal robustness.
\end{itemize}

\section{Related Work}

\subsection{Unimodal Adversarial Attack}
From an access perspective to the model, unimodal attacks can be divided into white-box and black-box attacks. 
In the black-box scenario, due to the target network's opaque weights, attackers typically conduct white-box attacks on an accessible source network, then transfer the adversarial examples to the target network. 
Therefore, the attack's transferability is also crucial.

\textbf{White-box Attack.}
Based on how to constrain perturbation, adversarial attacks on the visual modality can generally be divided into two categories.
(1) Global attacks typically involve perturbing all pixels of an image, usually constraining the distance between adversarial and original images based on \(\ell_\infty\), \(\ell_2\), or \(\ell_1\) norms. 
Representative methods include FGSM \cite{fgsm}, PGD \cite{pgd}, APGD \cite{apgd1}, CW \cite{cw}, etc. 
(2) Patch attacks, which confine the perturbation to a small area, such as 2\% of the image, and allow unrestricted modification of image pixels within that area. 
Representative methods include LaVAN \cite{lavan} and Depatch \cite{depatch}. 
Since patch attacks are more practical, physical world attacks are usually based on this form.
In the text domain, due to the discrete nature of text data, such attacks typically involve subtle modifications to the original text, such as replacing synonyms, inserting additional words, or adjusting sentence structure, without significantly altering the meaning of the text.
A representative method is BertAttack \cite{bertatk}.

\textbf{Boosting Transferability.} Enhancing the transferability of adversarial attacks is essentially a generalization problem. 
The two main approaches to solving the generalization issue are data augmentation and improving the optimization algorithm, thus dividing transfer attack methods into two categories.
(1) Typical methods that boost transferability through data augmentation, such as DI \cite{di} (Diverse Inputs), TI \cite{ti} (Translation Invariant) and SI \cite{ni} (Scale Invariant).
(2) Typical schemes that improve optimization algorithm, such as MI \cite{mi} (Momentum Iterative), NI \cite{ni} (Nesterov Iterative), VMI \cite{vmi} (Varied Momentum Iterative), VNI \cite{vmi} (Varied Nesterov Iterative).

\subsection{Multimodal Adversarial Attack}
This subsection discusses relevant VLP models and multimodal adversarial attack methods.

\textbf{VLP Models.}
VLP models based on different combinations of pre-training tasks can be roughly divided into three categories.
(1) Aligned models: CLIP \cite{clip} contains two unimodal encoders to align multimodal embeddings based on Image-Text Contrastive (ITC) loss.
(2) Fused models by matching: ViLT \cite{vilt} introduces both Image-Text Matching (ITM) and Masked Language Modeling (MLM) pre-training for V+L tasks. Models like ALBEF \cite{albef}, TCL \cite{tcl}, BLIP \cite{blip}, and VLMo \cite{vlmo} build on it, first aligning multimodal features using ITC loss, then fusing cross-modal features using ITM and MLM losses.
(3) Fused models by Masked Data Modeling (MDM): BEiT \cite{beit} and BEiTV2 \cite{beit2} propose and improve Masked Image Modeling (MIM) loss. BEiT3, based on it, first aligns multimodal features using ITC loss, then fuses cross-modal features using MIM, MLM, and Masked Language-Vision Modeling (MLVM) losses.

\textbf{Multimodal Attack.}
Attacking VLP models is a novel topic.
Existing work has provided valuable insights. 
Co-Attack \cite{coattack} designs general optimization objectives based on different embeddings (unimodal or multimodal) and experimentally demonstrates that using text attacks or image attacks alone is not as effective as using both in combination, providing a general baseline for subsequent works.
SGA \cite{sga} points out that improving the diversity of multimodal interaction can enhance the transferability of multimodal adversarial examples. 
AdvCLIP \cite{advclip} provides a framework to learn a universal adversarial patch on pre-trained models for transfer attacks on downstream fine-tuned models.
These works are limited to VLP models and ignore the connection between unimodal and multimodal scenarios, which is our main motivation.

\section{Methodology}
\subsection{Feature Guidance} \label{sec:Feature Guidance}
An image feature extractor \(E\) (e.g., an image contrastive representation encoder \cite{moco,simclr,byol} or a VLP model's visual encoder) projects the image into a feature vector for various visual tasks like image classification and object detection.
% We consider an image feature extractor $E$ (such as an image contrastive representation encoder  or the visual encoder in a VLP model) responsible for projecting the image into a feature vector, which might be fed into a subsequent structure to perform various visual downstream tasks such as image classification, object detection, or various V+L downstream tasks. 
Without regarding the subsequent usage, the most intuitive approach to generate an adversarial example $x'$ for an image $x$ is to encourage the feature vectors $E(x')$ and $E(x)$ to be as distant as possible \cite{coattack}.
% We refer to this universal strategy as \textbf{``Feature Deviation Attack" (FDA)} involving maximizing the loss function:
This universal strategy is termed “Feature Deviation Attack" (FDA), which involves maximizing the loss function:
\begin{equation}
L_{dev} = -E(x') \cdot E(x) .
\end{equation}
where \(\cdot\) represents the dot product of vectors, \(E(x)\in \mathbb{R}^{d}\).

Assuming that in the embedding space, there exists a set of guiding vectors \( W = \{\omega_i\}_{i=1}^{m} \), \(\omega \in \mathbb{R}^{d}\), and there is a set of guiding labels \( Y = \{y_i\}_{i=1}^{n} \), \(y\in \{1,2,\ldots,m\}\) specifying that \( E(x') \) should be distant from the guiding vectors \( \{\omega_{y_i}\}_{i=1}^{n} \in W \).
We refer to this strategy as \textbf{``Feature Guidance Attack" (FGA)}. 
To realize the above concept, we need to maximize the loss function:
\begin{equation}
 L_{gui} = -\frac{1}{n} \sum_{i=1}^{n} ln\left( \frac{\exp(E(x') \cdot \omega_{y_i})}{\sum_{j=1}^{m}\exp(E(x') \cdot \omega_{j})} \right) 
\end{equation}
where \(exp(\cdot)\) represents the exponential function with Euler's number \(e\) as the base, and \(ln(\cdot)\) stands for the logarithm to the base \(e\).

Based on \(L_{gui}\) or \(L_{dev}\), we can apply the PGD process \cite{pgd}, gradually pushing the clean example \( x \) along the gradient direction to maximize the loss function, ultimately obtaining the adversarial example \( x' \).
By the chain rule of gradients, \( \frac{\partial L}{\partial x'} = \frac{\partial L}{\partial E(x')} \cdot \frac{\partial E(x')}{\partial x'} \). 
\( \frac{\partial L}{\partial x'} \) represents the direction of perturbation added to the input example. 
While we focus on \( \frac{\partial L}{\partial E(x')} \) which represents the movement direction of the feature vector:
\begin{equation}
\frac{\partial L_{dev} }{\partial E(x')}=-E(x)
\end{equation}
\begin{equation}
\frac{\partial L_{gui} }{\partial E(x')}= \sum_{i=1}^{n}\left( -\frac{1}{n}\cdot\omega_{y_i}\right)+\sum_{k=1}^{m} \frac{exp(E(x')\cdot \omega_{k})}{\sum_{j=1}^{m}exp(E(x')\cdot \omega_{j} ) } \cdot \omega _k
\end{equation}

It can be observed that feature deviation loss promotes the movement of \( E(x') \) towards \( -E(x) \), which means moving away from \( E(x) \). 
While, Regarding the first term of \(\partial L_{gui} /\partial E(x')\), it encourages \( E(x') \) to move away from the guiding vectors \( \{\omega_{y_i}\}_{i=1}^{n} \), and assigning equal weight \( 1/n \) to each of them.
The second term encourages \( E(x') \) to approach the guiding vector \( \omega_k \in W\), with a weight of \( \exp(E(x') \cdot \omega_k)/\sum_{j=1}^{m}\exp(E(x') \cdot \omega_j) \), which means the closer \( E(x') \) is to a guiding vector, the greater the weight assigned to it.
Due to the presence of the first term, \( E(x') \) is far from \( \{\omega_{y_i}\}_{i=1}^{n} \), resulting in the weight of \( \omega_{y_i} \) being almost zero in the second term. 
Consequently, the second term effectively facilitates \( E(x') \) in selecting a nearby guiding vector that does not belong to the set \( \{\omega_{y_i}\}_{i=1}^{n} \) and moving closer to it.

% \begin{table}[ht]
% \caption{Attacking results of SimCLR encoder on CIFAR-10. The reported value is classification accuracy.}
% \label{tab:classification}
% \begin{tabular}{ccc}
% \toprule
% Method & Top1  & Top5  \\
% \midrule
% without attack & 88.09 & 98.53 \\
% deviation attack & 12.38 & 53.67 \\
% guidance attack & \textbf{1.70}  & \textbf{33.69} \\
% \bottomrule
% \end{tabular}
% \end{table}
\begin{figure}[t]
  \centering
  \includegraphics[width=\linewidth]{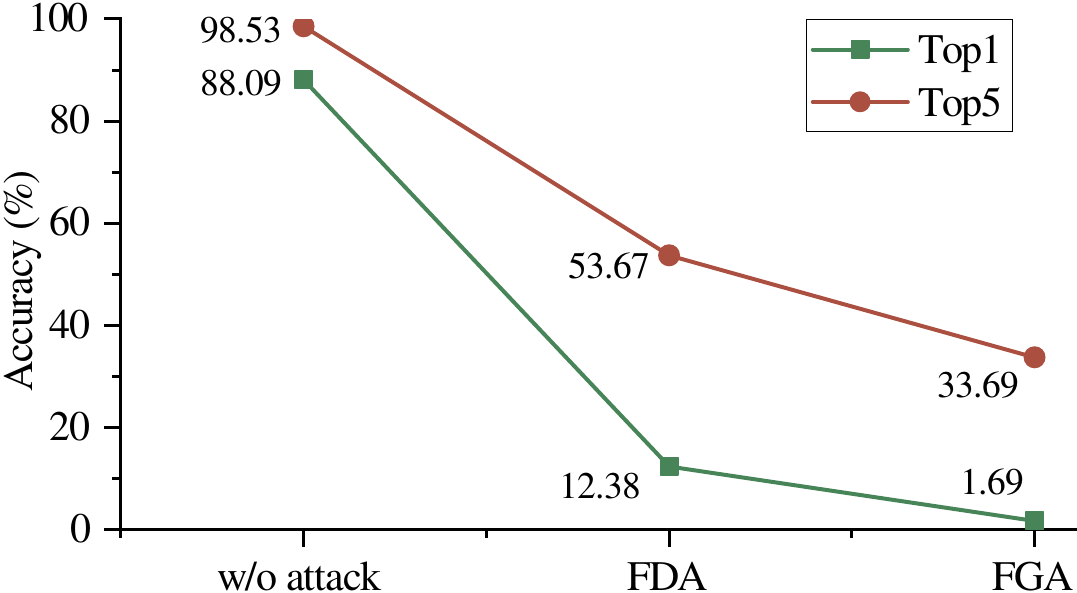}
  \caption{Attacking results of SimCLR encoder on CIFAR-10. The reported value is classification accuracy.}
  \label{fig:classification}
% Adversarial image is achieved by guiding its embedding away from the aligned space into a non-aligned space, while ensuring the embedding does not approach the adversarial text embeddings.}
  % \vspace*{\fill}
  \Description{Attacking results of SimCLR encoder on CIFAR-10. The reported value is classification accuracy.}
\end{figure}
% \vspace*{\fill}

We conduct a simple attack experiment using the SimCLR image encoder \cite{simclr} and the CIFAR-10 dataset \cite{cifar10}, where image feature vectors are used for image classification through a KNN-200 classifier. 
During the feature guidance attack, we first use the encoder to extract features for all training data. 
Then, by averaging the features belonging to the same category, we obtain ten guiding vectors \( \{\omega_1, \omega_2, \ldots, \omega_{10}\} \). 
The label \( y \in \{1,2, \ldots, 10\} \) of the image \( x \) serves as the guiding label, encouraging \( E(x') \) to move away from the guiding vector \( \omega_y \).
From Table~\ref{fig:classification}, it is not difficult to observe that the intensity of the feature guidance attack is greater than the feature deviation attack.

Most existing VLP models typically consist of two unimodal encoders, a text encoder \( E_t \) and a visual encoder \( E_v \), along with a multimodal feature fusion encoder \( E_m \). 
For a single image-text paired example \( (v, t) \), it is first mapped to a shared feature space separately by \( E_t \) and \( E_v \) for aligning image and text features. 
Subsequently, cross-modal feature fusion is conducted through \( E_m \). 
% Therefore, in VLP models, there are three key embeddings of interest: \( E_v(v) \), \( E_t(t) \), and \( E_m(E_v(v), E_t(t)) \), noting that all refer to the [CLS] vector. 
Therefore, VLP models focus on three key embeddings:\( E_v(v) \), \( E_t(t) \), and \( E_m(E_v(v), E_t(t)) \), all corresponding to the [CLS] vector.
We will focus on finding guiding vectors in the embedding space and constructing guiding labels to execute FGA on VLP models.

\subsection{Attacking after Fuse} 
In this scenario, we focus on the fused embedding \( E_m(E_v(v), E_t(t)) \). 
For different V+L downstream tasks, it needs to be fed into different subsequent models which can be uniformly understood as comprising a projector $P$ and a linear classification head $h$.
\(P\) projects the fused embedding into a task-specific downstream embedding space, followed by \(h\) performing classification on this embedding.
% $P$ is responsible for projecting the fused embedding into a task-specific downstream embedding space, and then $h$ performs classification on the projected embedding.
% Taking Visual Entailment (VE) as an example, the fused embedding is fed into a MLP to obtain classification results. 
% In this setup, the input layer and all hidden layers of the MLP constitute $P$, projecting the fused embedding into the VE embedding space. 
% The output layer of the MLP acts as $h$, tasked with determining whether the \( (v, t) \) pair is contradiction, neutral, or entailment.
% Similarly, this logic applies to other downstream tasks such as Visual Question Answering (VQA) and Visual Reasoning (VR). 
We can rewrite \( P(E_m(E_v(v), E_t(t))) \) as \( E(v|t) \), obtaining an image encoder conditioned on the textual modality. 
The weight of the linear classification head, \( W = \{\omega_i\}_{i=1}^{c} \in \mathbb{R}^{d \times c} \) ($c$ is the number of categories), serve as guiding vectors. 
Using label information as guiding labels \( Y \), we thus have all the necessary components to implement the Feature Guidance Attack.
See Appendix A for more details and explanations about Visual Question Answering (VQA), Visual Reasoning (VR), etc. 

FGA is primarily used to generate image adversarial examples. 
However, for VLP models, attacking both modalities simultaneously is a more effective strategy \cite{coattack,sga}. 
The main challenge in generating adversarial texts involves solving the following optimization problem:
\begin{equation} 
\label{equ:txtattackafterfuse}
    t'=\underset{t'}{argmax} (\left \| E_m(E_v(v),E_t(t')) -E_m(E_v(v),E_t(t))\right \| )
\end{equation}
where BertAttack \cite{bertatk} is a well-suited choice for addressing this problem.

In fact, FGA and BertAttack are completely orthogonal strategies. 
This means we first generate an adversarial text example \( t' \) and then use \( E(v|t') \) as the image encoder to perform FGA, obtaining \( v' \). 
Thus, we acquire the adversarial pair \( (v', t') \).

\begin{figure}[t]
  \centering
  \includegraphics[width=\linewidth]{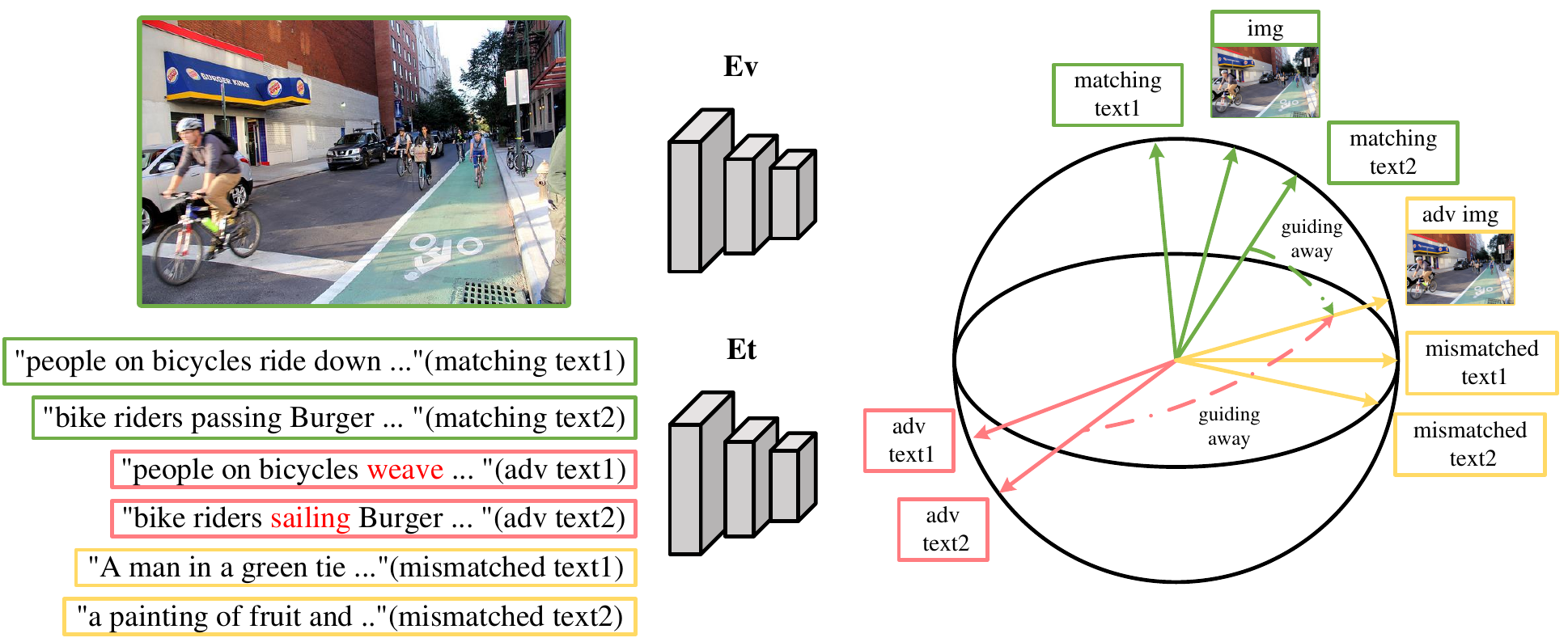}
  \caption{Illustration of Feature Guidance with Text Attack (FGA-T) before fuse.}
% Adversarial image is achieved by guiding its embedding away from the aligned space into a non-aligned space, while ensuring the embedding does not approach the adversarial text embeddings.}
  % \vspace*{\fill}
\Description{Illustration of Feature Guidance with Text Attack (FGA-T) before fuse.}
\end{figure}
% \vspace*{\fill}

\subsection{Attacking before Fuse}
% In this scenario, we only use two unimodal encoders: \( E_v \), and \( E_t \).
In this scenario, only two unimodal encoders are used: \( E_v \), and \( E_t \).
The primary intention of VLP models is to learn directly from raw text descriptions of images, utilizing a broader source of supervision \cite{clip}. 
This implies that in CV models, image understanding comes from pre-provided labels, such as image categories, pixel categories, or annotated bounding boxes.
% This suggests that in CV models, the understanding of images is derived from pre-provided label information, such as image categories, pixel categories, or annotated bounding boxes. 
In contrast, in VLP models, the understanding of images originates from raw text. 
% Hence, a natural idea emerges to use text as supervisory information for generating image adversarial examples.
Consequently, using text as supervisory information to generate image adversarial examples becomes a natural approach.
To implement this approach, we first acquire a text set \( T = \{t_i\}_{i=1}^{m} \). 
Then, We use the text encoder to obtain a set of guiding vectors \( \{\omega_i\}_{i=1}^{m} = \{E_t(t_i)\}_{i=1}^{m} \). 
To obtain this text set $T$, all texts are gathered from the dataset. 
Here, by utilizing the dataset's annotations, we can identify which texts in the text set match with the image \( v \), thereby obtaining the guiding labels \( Y \). 
% For the second source, we use a predefined set of nouns (such as the 1000 category names from the ImageNet dataset[]) to construct a series of texts, like "there is a [xxx] in this photo."
% Subsequently, we compare the cosine similarity between \( E_v(v) \) and \( E_t(t_i) \), and select the top \( n \) texts with the highest similarity to construct the guiding labels \( Y =\{y_1,y_2,\ldots,y_n\}\):
% \begin{equation}
% y_k=\underset{i}{argmaxtopk} \left(\frac{E_v(v)\cdot E_t(t_i)}{\left \| E_v(v) \right \|_2 \cdot\left \|E_t(t_i) \right \|_2 } \right)
% \end{equation}
By this point, all elements necessary for executing FGA have been acquired: the image encoder \( E_v \), the set of guiding vectors \( \{E_t(t_i)\}_{i=1}^{m}  \), and the guiding labels \( Y \).
By maximizing \(L_{gui}\), the feature vector \(E_v(v')\) will diverge from the text representations \(\{E_t(t_y)\}_{y \in Y}\) that match \(v\), thereby generating adversarial images \(v'\). 
For more details on executing iterations of FGA and how it can be combined with typical attack strategies in the unimodal domain, refer to Appendix B.

% We will validate the effectiveness of the attacks introduced in this subsection based on two tasks: Image-Text Retrieval (ITR) and Zero-Shot Classification (ZC).

% Experimental evidence shows that under white-box condition before fuse, image modality perturbation solely through FGA is sufficient to incapacitate VLP models. 
% Consequently, we temporarily don't consider introducing text attack.
% In the next subsection, we will continue to introduce how to incorporate text attack before fuse and combine other strategies to enhance the transferability of adversarial examples in black-box scenario.

\subsection{Boosting Transferability before Fuse}
SGA\cite{sga} points out that multimodal adversarial examples have better transferability than unimodal adversarial examples. 
Therefore, we need to introduce text attack into FGA before fuse.
We consider an image minibatch \( V = \{v_i\}^n_{i=1} \) and the text set \(T_i\) represents all texts that match with the image \( v_i \).
Firstly, for each text \( t \in T_i \), we handle the following optimization problem to generate adversarial text \( t' \):
\begin{equation}
  t'=\underset{t'}{argmax} \left(-\frac{E_t(t')\cdot E_v(v_i)}{\left \| E_t(t') \right \|_2\left \| E_v(v_i) \right \|_2  } \right)
\end{equation}
where \(\left \| \cdot\right \|_2\) denotes the Euclidean distance.

At this point, we obtain the adversarial text set \( T_i' \), and we denote \( T = T_1 \cup T_2  \ldots \cup T_n \cup T_1' \cup T_2' \ldots \cup T_n' \).
Secondly, for \( v_i \in V \), where \( T_i \) is its matching texts and \( T_i' \) is the adversarial texts, to generate adversarial example \( v_i' \), we use the feature guidance loss to encourage \( E_v(v_i') \) to simultaneously move away from both \( E_t(T_i) \) and \( E_t(T_i') \):
\begin{equation}
L_{gui}(v_i') = -\frac{1}{len_i} \sum_{t^*\in T_i \cup T_i'} log\left( \frac{\exp(E_v(v_i') \cdot E_t(t^*))}{\sum_{t\in T}\exp(E_v(v_i') \cdot E_t(t))} \right) 
\end{equation}
\( len_i \) represents the length of text set \( T_i \cup T_i' \).

Building on this foundation, we further introduce two strategies to enhance transferability:
(1) Following SGA \cite{sga}, we preset a set of resize parameters \( S = \{s_1, s_2, \ldots, s_m\} \), where \( h(v, s_k) \) denotes the resizing function that takes the image \( v \) and the scale coefficient \( s_k \) as inputs.
After data augmentation, the objective function we aim to maximize is no longer \( L_{gui}(v_i') \) but rather \( \sum_{k=1}^{m} L_{gui}(h(v_i', s_k)) \), where \( h(v_i', s_k) \) represents the augmented image.
(2) Following MI-FGSM \cite{mi}, we introduce the momentum mechanism, where the current perturbation direction is determined by both the current gradient and the historical gradients from previous iterations. 
See Appendix B.2 for more details.

\section{Experiments}
\subsection{Experimental Setting}
\subsubsection{VLP Models}
Our experimental section involves four typical VLP models: CLIP, ALBEF, TCL and BEiT3. 
CLIP is a typical aligned model, consisting solely of two unimodal encoders. 
The latter three are fused models, containing two unimodal encoders and a multimodal encoder. 
ALBEF and TCL share the same architecture with some differences in the details of ITC loss.
Besides, ALBEF and BEiT3 have two main differences: (1)
\textbf{Different Pre-training Tasks:} ALBEF is based on three pre-training tasks: ITC, ITM and MLM. 
In contrast, BEiT3 is based on three MDM tasks: MLM, MIM, and MVLM.
(2)
\textbf{Different Model Structures:} In ALBEF, the three encoders are independent of each other. 
BEiT3, however, uses the Multiway Transformer to split the feed-forward layer into three parallel paths, thereby obtaining three encoders.

\subsubsection{V+L Downstream Tasks}
In this part, we will introduce each downstream task involved in the experiments, along with the models and datasets used to perform these tasks.

\textbf{Visual Entailment (VE)} is a fine-grained visual reasoning task, where given a pair of \((v,t)\), the model needs to determine whether the text is supported by the image (entailment), whether the text is unrelated to the image (neutral), or whether the text contradicts the content of the image (contradictory).
This task will be conducted based on the ALBEF model and the SNLI-VE \cite{snlive} dataset.

\textbf{Visual Question Answering (VQA)} requires the model to predict a correct answer given an image and a question \cite{vqa1, vqa2}.
It can be viewed as a multi-answer classification problem, or as an answer generation problem. 
We use the VQAv2 \cite{vqa1} dataset and the ALBEF model, which performs the VQA task through text generation.

\textbf{Visual Grounding (VG)} requires the model to find parts of the image that match the given textual description.
We perform this task based on the RefCOCO+ \cite{refcoco} dataset and the ALBEF model.

\textbf{Visual Reasoning (VR)} requires the model to predict if a given text describes a pair of images. 
This task necessitates that the model not only understands the content of individual images but also compares and reasons about the relationship between two images. 
Therefore, the input consists of a pair of images and a piece of text. 
We use the BEiT3 model and the NLVR2 \cite{nlvr2} dataset to perform this task.

\textbf{Zero-Shot Classification (ZC)} requires using predefined category descriptions (such as "a cat," "a car," etc.) as text inputs and mapping these descriptions to the embedding space by text encoder. 
Then, for a given image, the similarity between the image embedding and each category description embedding is calculated, and the image is classified into the category with the highest similarity.
Due to the CLIP model's strong zero-shot capacity, we use it along with three datasets: CIFAR-10 \cite{cifar10}, CIFAR-100 \cite{cifar10}, and ImageNet \cite{imagenet}, to perform this task.

\textbf{Image-Text Retrieval (ITR)} involves retrieving relevant images from an image database given a text query, and vice versa \cite{itr1, itr2, itr3}.
We perform this task based on the CLIP, ALBEF and TCL models, and the Flickr30k \cite{flickr30k} and MS COCO \cite{coco} datasets.

% \begin{table}[h]
%   \begin{tabular}{lllllll}
%            & VE      & VQA   & VG       & VR    & ZC                                                                        & ITR                \\
%   Models   & ALBEF   & ALBEF & ALBEF    & BEiT3 & CLIP                                                                      & CLIP, ALBEF, TCL   \\
%   Datasets & SNLI-VE & VQAv2 & RefCOCO+ & NLVR2 & \begin{tabular}[c]{@{}l@{}}CIFAR10, \\ CIFAR100, \\ ImageNet\end{tabular} & Flickr30k, MS COCO
% \end{tabular}
% \end{table}

\begin{table}[t]
\caption{Comparison results on four downstream tasks after fuse. The reported value is accuracy. Lower is better.}
\label{tab:afterfuse}
\setlength{\tabcolsep}{2.8pt}
\begin{tabular}{l|cccccccc}
\toprule
\multicolumn{1}{c|}{\multirow{2}{*}{Method}} & VE    & \multicolumn{2}{c}{VQA} & \multicolumn{3}{c}{VG} & \multicolumn{2}{c}{VR} \\
\cmidrule(r){2-2} \cmidrule(r){3-4} \cmidrule(r){5-7} \cmidrule(r){8-9} 
\multicolumn{1}{c|}{}  & test  & dev   & std   & val    & testA & testB & dev         & test-P      \\ 
\midrule
w/o atk    & 79.91 & 75.83      & 76.04      & 58.44  & 65.91 & 46.25 & 83.54       & 84.38       \\
\midrule
TA   & 55.09 & 45.47      & 45.89      & 49.17  & 54.05 & 39.27 & 69.59       & 70.46       \\
IA            & 42.72 &   52.78   &     52.88   & 45.78  & 51.48 & 36.16 & 63.10       & 63.14       \\
SA            & 38.42 & 41.21   & 41.31       & 42.13  & 45.93 & 34.96 & 58.43       & 58.53       \\
CA           & 19.36 & 36.91  & 37.01   & 36.61  & 39.87 & 30.21 & 54.77       & 54.67       \\
\midrule
FGA           & 5.66  & 48.70      & 48.77      & 36.54  & 42.18 & 29.33 & 0.93        & 1.15        \\
FGA-T           & \textbf{2.78}  &  \textbf{35.46}   &    \textbf{35.70}  & \textbf{34.11} & \textbf{38.16} & \textbf{28.86} & \textbf{0.52}        & \textbf{0.70}        \\
FGA$^1$       & 39.05 & 60.66      & 60.65      & 41.68  & 45.09 & 36.41 & 27.60       & 28.79       \\
FGA-T$^1$      & 22.37 & 41.00      &  41.07  & 35.54  & 39.38 & 30.44 & 19.15       & 20.49       \\
FGA$_{\ell_1}$  & 8.26  &  53.47 &   53.54  & 38.70   &    44.88   &   30.76    & 1.46      & 1.74        \\
FGA$_{pat}$   & 5.23  &  51.24  &   51.22   &   55.92     &   64.23    &    45.26   & 7.59      & 8.43       \\
\bottomrule
\end{tabular}
\end{table}

\subsection{Attack Effectiveness after Fuse}

This subsection explores the effectiveness of attacks on the fused feature vector.
Since VE, VQA, VG, and VR rely on this vector, we choose to evaluate these four tasks. 
As Table~\ref{tab:afterfuse} illustrates, the evaluation follows the baseline set by \cite{coattack},
\textbf{TA} represents alone text attack using BertAttack.
\textbf{IA} represents image attack based on feature deviation loss.
\textbf{SA} stands for separate unimodal attack, indicating that TA and IA are executed separately without modal interaction, and \textbf{CA} denotes the multimodal white-box attack Co-Attack \cite{coattack} which introduces cross-modal interaction.
For fairness, we set the \(\ell_{\infty}\) perturbation constraint for the image modality in FGA and FGA-T to \(\epsilon = 2/255\) with 10 iterations, consistent with IA, SA, and CA. 
Additionally, we explore the attack effectiveness when the number of iterations is 1, i.e., single-step attack, namely FGA$^1$ and FGA-T$^1$. 
We also investigate the effectiveness under the \(\ell_1\) constraint, namely FGAl$_{\ell_1}$, with \(\epsilon_{\ell_{1}} = 255\) and 20 iterations.
(See Appendix B.1 for more details.)
Besides, we perform FGA in patch form, namely FGA$_{pat}$, with 100 iterations, a single-step $\ell_{\infty}$ constraint of \(\alpha = 8/255\), and a patch area of 2\% of the total image area, with a random location. 
(See Appendix B.3 for more details.)
Furthermore, when involving text attack, BertAttack is used with a restriction of 1 perturbable token, following \cite{coattack,sga}.

We can observe from Table~\ref{tab:afterfuse}: 
(1) Under all tasks, FGA-T consistently achieves the best white-box attack performance, validating the effectiveness of the feature guidance approach and its orthogonality with text attack.
(2) Even with only a single step, the feature guidance method is sufficient to produce effective adversarial examples, performing on par with or even better than the baseline. 
This provides a faster and more convenient attack strategy.
(3) The feature guidance approach exhibits good orthogonality with other attack strategies in Computer Vision. When combined with \(\ell_1\) attack or patch attack, it demonstrates strong performance.

\begin{table}[t]
\caption{Comparison results on ZC task before fuse on CLIP. The reported value is accuracy. Lower is better.}
\label{tab:zeroclass}
\begin{tabular}{c|l|ccc}
\toprule
Metric                & Method          & CIFAR-10 & CIFAR-100 & ImageNet\\
\midrule
\multirow{6}{*}{Top1} & w/o atk      & 89.24   & 64.76    & 62.308   \\
                      & IA              & 35.5    & 17.04    & 15.29    \\
                      & FGA             & \textbf{0.01}    & \textbf{0.0}      & \textbf{0.004}    \\
                      & FGA\(^1\)       & 30.10   & 12.21    & 5.46     \\
                      & FGA\(_{\ell_1}\)& 14.5    & 7.06     & 0.028    \\
                      & FGA\(_{pat}\) & 0.1     & 0.0      & 0.01     \\
\midrule
\multirow{6}{*}{Top5} & w/o atk      & 98.94   & 86.9     & 86.78    \\
                      & IA              & 78.21   & 34.89    & 31.36    \\
                      & FGA             & 10.54   & 0.56     & \textbf{0.468}    \\
                      & FGA\(^1\)       & 79.26   & 34.39    & 27.12    \\
                      & FGA\(_{\ell_1}\)& 24.59   & 9.06     & 0.506    \\
                      & FGA\(_{pat}\) & \textbf{9.66}    & \textbf{0.35}     & 0.708   \\
\bottomrule
\end{tabular}
\end{table}

\begin{table*}[t]
\caption{Comparison results on image-text retrieval before fuse on CLIP.  For text-retrieval (TR) and
image-retrieval (IR), R@1, R@5 and R@10 are reported respectively. Lower is better.}
\label{tab:itrclip}
\setlength{\tabcolsep}{6pt}
\begin{tabular}{l|l|ccc|ccc|ccc|ccc}
\toprule
\multicolumn{2}{c|}{\multirow{3}{*}{Method}} & \multicolumn{6}{c|}{Flickr30k(1K test set)}                         & \multicolumn{6}{c}{MSCOCO(5K test set)}                        \\
\multicolumn{2}{c|}{}                        & \multicolumn{3}{c}{TR} & \multicolumn{3}{c|}{IR} & \multicolumn{3}{c}{TR} & \multicolumn{3}{c}{IR} \\
\cmidrule(r){3-5} \cmidrule(r){6-8} \cmidrule(r){9-11} \cmidrule(r){12-14} 
\multicolumn{2}{c|}{}                        & R@1     & R@5    & R@10   & R@1     & R@5    & R@10   & R@1     & R@5    & R@10   & R@1     & R@5    & R@10   \\
\midrule
\multicolumn{2}{c|}{w/o attack}                      & 81.5   & 96.3  & 98.4  & 62.08  & 85.62 & 91.7  & 52.42  & 76.44 & 84.42 & 33.02  & 58.16 & 68.4  \\
\midrule
\multirow{4}{*}{\parbox{1.5cm}{Feature \\Deviation}} & TA          & 61.8   & 85.8  & 92.0  & 41.18  & 66.68 & 76.78 & 27.42  & 51.06 & 62.54 & 16.40  & 34.49 & 44.52 \\
                            & IA          & 25.6   & 47.6  & 56.4  & 19.84  & 39.18 & 48.94 & 10.84  & 24.16 & 32.06 & 6.76   & 17.76 & 24.71 \\
                            & SA          & 17.2   & 35.7  & 45.9  & 11.14  & 25.32 & 33.32 & 5.8    & 14.22 & 19.8  & 3.21   & 9.39  & 14.06 \\
                            & CA          & 7.3    & 16.5  & 22.4  & 4.18   & 10.04 & 13.86 & 1.6    & 4.62  & 7.08  & 0.94   & 2.89  & 4.53  \\
\midrule
\multirow{2}{*}{\parbox{1.5cm}{MS COCO \\Categories}}          & FGA\(^1\)         & 68.9   & 89.4  & 93.5  & 49.66  & 75.08 & 83.48 & 40.46  & 64.54 & 74.08 & 23.89  & 46.57 & 57.68 \\
                            & FGA               & 16.6   & 32.2  & 39.6  & 11.54  & 25.78 & 33.68 & 5.02   & 11.06 & 14.78 & 2.76   & 7.37  & 10.43 \\
\midrule
\multirow{2}{*}{\parbox{1.5cm}{ImageNet\\Categories} }        & FGA\(^1\)         & 66.8   & 87.5  & 92.5  & 48.48  & 73.94 & 82.66 &  39.80  & 63.03  &  73.04  & 23.67   &   46.07  &  57.33  \\
                            & FGA               & 11.5   & 21.0  & 28.3  & 7.64   & 17.8  & 23.1  &   3.54 &  8.28  &  11.44  &   2.25  &   6.02  &  8.68  \\
\midrule
\multirow{4}{*}{Test Texts}  & FGA\(^1\)          & 27.5   & 49.1  & 59.5  & 17.72  & 38.92 & 50.26 & 14.54  & 30.4  & 39.88 & 8.18   & 21.32 & 30.18 \\
                            & FGA                & 0.0    & 0.8   & 1.6   & 0.14   & 0.44  & 0.96  & 0.06   & 0.24  & 0.4   & 0.024  & 0.152 & 0.264 \\
                            & FGA\(_{\ell_1}\)   & 0.1    & 0.2   & 0.5   & 0.12   & 0.32  & 0.5   & 0.04   & 0.12  & 0.16  & 0.068  & 0.200 & 0.280 \\
                            & FGA\(_{pat}\)    & 0.2    & 0.4   & 0.4   & 0.18   & 0.48  & 0.78  & 0.08   & 0.16  & 0.24  & 0.080  & 0.200 & 0.312 \\
\bottomrule
\end{tabular}
\end{table*}

\begin{table*}[t]
\caption{Comparison results on image-text retrieval before fuse on ALBEF. For text-retrieval (TR) and
image-retrieval (IR), R@1, R@5 and R@10 are reported respectively. Lower is better.}
\label{tab:itralbef}
\setlength{\tabcolsep}{6pt}
\begin{tabular}{l|l|ccc|ccc|ccc|ccc}
\toprule
\multicolumn{2}{c|}{\multirow{3}{*}{Method}} & \multicolumn{6}{c|}{Flickr30k(1K test set)}                         & \multicolumn{6}{c}{MSCOCO(5K test set)}                        \\
\multicolumn{2}{c|}{}                        & \multicolumn{3}{c}{TR} & \multicolumn{3}{c|}{IR} & \multicolumn{3}{c}{TR} & \multicolumn{3}{c}{IR} \\
\cmidrule(r){3-5} \cmidrule(r){6-8} \cmidrule(r){9-11} \cmidrule(r){12-14} 
\multicolumn{2}{c|}{}                        & R@1     & R@5    & R@10   & R@1     & R@5    & R@10   & R@1     & R@5    & R@10   & R@1     & R@5    & R@10   \\
\midrule
\multicolumn{2}{c|}{w/o attack}   & 95.9 & 99.8 & 100.0 & 85.5  & 97.5  & 98.9  & 77.58 & 94.26 & 97.16 & 60.67 & 84.33 & 90.51 \\
\midrule
\multirow{4}{*}{\parbox{1.5cm}{Feature \\Deviation}} & TA          & 85.8 & 98.1 & 98.9  & 64.1  & 83.68 & 88.16 & 53.08 & 78.32 & 86.7  & 34.48 & 59.38 & 69.08 \\
                                                    & IA          & 47.4 & 65.6 & 71.4  & 38.64 & 56.74 & 62.82 & 30.26 & 47.7  & 55.5  & 21.19 & 38.16 & 46.05 \\
                                                    & SA          & 31.6 & 50.6 & 58.4  & 23.66 & 39.68 & 46.64 & 15.44 & 29.54 & 36.74 & 10.21 & 21.89 & 28.27 \\
                                                    & CA          & 32.5 & 50.9 & 58.4  & 23.42 & 39.5  & 45.92 & 14.58 & 28.26 & 35.5  & 9.90  & 21.78 & 27.92 \\
\midrule
\multirow{2}{*}{Test Texts}  & FGA\(^1\)          & 38.0 & 58.0 & 65.3  & 31.26 & 52.12 & 60.72 & 27.96 & 48.76 & 57.88 & 20.79 & 41.26 & 51.40 \\
                            & FGA                & \textbf{0.7}  & \textbf{1.0}  & \textbf{1.1 }  & \textbf{0.54}  & \textbf{0.94}  & \textbf{1.1}   & \textbf{0.32}  & \textbf{0.78}  & \textbf{1.02}  &\textbf{ 0.27}  &\textbf{ 0.76}  & \textbf{1.12} \\    
\bottomrule
\end{tabular}
\end{table*}

\subsection{Attack Effectiveness before Fuse}

In this subsection, we explore the effectiveness of attacks on two unimodal encoders. 
The tasks of ZC and ITR primarily rely on two unimodal embeddings. 
We first conduct attacks on the ZC task. 
Since the text input in the ZC task is predefined and cannot be altered, we only use attacks involving the visual modality. 
When conducting FGA, we construct the text ``A photo of a \{object\}.'' using the categorys' name to obtain the text set \( T = \{t_i\}_{i=1}^c \), where \( c \) represents the number of categories, following \cite{clip}.
We extract features through \( E_t \) to construct the guiding vectors \( \{E_t(t_i)\}_{i=1}^c \), and the true category of the image serves as the guiding label \( y \in \{1,2,\ldots,c\} \).
The attack results are presented in the Table~\ref{tab:zeroclass}. 
We observe that even FGA$^1$ outperforms IA which is an iterative feature deviation attack.

When executing the ITR task on the CLIP model with ViT image encoder, to construct guiding vectors for FGA, we use not only all the texts in the dataset to construct the text set (``Test Texts'' in Table~\ref{tab:itrclip}), but also follow the approach of the ZC task: using the 1000 category names from the ImageNet dataset (``ImageNet Categories'' in Table~\ref{tab:itrclip}) or the 80 category names from the MS COCO dataset's object detection task (``MS COCO Categories'' in Table~\ref{tab:itrclip}) to construct texts ``There is a \{object\} in this photo.'' to form the text set \(T= \{t_i\}_{i=1}^c \).
Since in the Flickr30k and MS COCO datasets, an image may contain multiple objects, it is possible that the image matches multiple texts in \( \{t_i\}_{i=1}^c \). 
In fact, we do not have annotation information indicating which objects are in the image. 
Therefore, we compare the cosine similarity between \( E_v(v) \) and \( \{{E_t(t_i)}\}_{i=1}^c \) to find the top 5 texts with the highest cosine similarity to \( v \). 
When performing FGA, we encourage \( E_v(v') \) to move away from the feature vectors of these five texts.
From Table~\ref{tab:itrclip}, we can summarize:
(1) When using all texts to construct the feature guidance vectors, FGA achieves the best attack effect, which is intuitive. 
Moreover, we find that without the text attack, CLIP is already incapacitated on the ITR task.
(2) ImageNet includes more categories and therefore contains richer guiding information, resulting in better attack effects compared to using categories from COCO.

Since the CLIP model only contains two unimodal encoders, attacking before fuse actually utilizes the entire CLIP model. 
However, the ALBEF model additionally includes a multimodal encoder, so attacking before fuse ignores the multimodal encoder. 
Therefore, it is necessary to validate the effectiveness of FGA before fusion on the ALBEF model. 
As shown in Table~\ref{tab:itralbef}, we conduct this experiment based on the ALBEF model and the ITR task and observe phenomena consistent with Table~\ref{tab:itrclip}.

\subsection{Boosting Transferability}

In this subsection, we transition the attack from the white-box setting to the black-box setting, which is a more common scenario. 
We use four VLP models: ALBEF, TCL, CLIP\(_{ViT}\), and CLIP\(_{CNN}\). 
The TCL model is identical to ALBEF except for differences in the design of the Image-Text Contrastive (ITC) loss during training, resulting in different final network weights. 
The two CLIP models use ViT and CNN as visual encoders, respectively. 
The degree of difference between these four models varies, which will inevitably affect the transferability of adversarial examples. 
We will observe this phenomenon in the experiments. 
Our experimental setup is as follows:
% \begin{itemize}
%   \item \textbf{Task and Dataset:} We conduct black-box adversarial example transfer attacks based on the Image-Text Retrieval (ITR) task and the Flickr30k dataset.
%   \item \textbf{Source Model and Target Model:} The source model is the model for which we generate adversarial examples through white-box attacks, and then use them to attack the target model. 
%   Each model will serve as both source and target models.
%   \item \textbf{Attack Methods:} The methods we use involve attacking both image and text. 
%   SA and CA, which do not focus on transferability, serve as baselines. 
%   SGA is the state-of-the-art (SOTA) transfer attack and serves as the comparative method. 
%   FGA-T\(_{aug}\) is based on FGA-T with additional data augmentation using a set of resize parameters \( S  \), following SGA. 
%   The differences between SGA and FGA-T\(_{aug}\) are in the loss function used for generating adversarial images and the attack process (the former's attack order is ``text, image, text'', while the latter's attack order is ``text, image''). 
%   MFGA-T\(_{aug}\) additionally introduces the momentum mechanism.
%   \item \textbf{Hyperparameters:} All texts are allowed to modify only one word, all image perturbations are limited to 2/255 (\(\ell_{\infty}\) norm), and the number of iterations is 10, following \cite{coattack}. 
%   The resize parameters \( S = \{0.5, 0.75, 1.25, 1.5\} \), following SGA.
% \end{itemize}
(1) \textbf{Task and Dataset:} We conduct black-box adversarial example transfer attacks based on the Image-Text Retrieval (ITR) task and the Flickr30k dataset.
(2) \textbf{Source Model and Target Model:} The source model is the model for which we generate adversarial examples through white-box attacks, and then use them to attack the target model. 
Each model will serve as both source and target models.
(3) \textbf{Attack Methods:} The methods we use involve attacking both image and text. 
SA and CA, which do not focus on transferability, serve as baselines. 
SGA is the state-of-the-art (SOTA) transfer attack and serves as the comparative method. 
FGA-T\(_{aug}\) is based on FGA-T with additional data augmentation using a set of resize parameters \( S  \), following SGA. 
The differences between SGA and FGA-T\(_{aug}\) are in the loss function used for generating adversarial images and the attack process (the former's attack order is ``text, image, text'', while the latter's attack order is ``text, image''). 
MFGA-T\(_{aug}\) additionally introduces the momentum mechanism.
(4) \textbf{Hyperparameters:} All texts are allowed to modify only one word, all image perturbations are limited to 2/255 (\(\ell_{\infty}\) norm), and the number of iterations is 10, following \cite{coattack}. 
The resize parameters \( S = \{0.5, 0.75, 1.25, 1.5\} \), following SGA.

\begin{table*}[t]
\caption{Compare the transferability with SOTA methods based on the Flickr30k dataset. The reported value is the attack success rate. Higher is better. R@1 value after the attack is reported in parentheses for SGA, FGA-T$_{aug}$, and MFGA-T$_{aug}$. Lower is better.}
\label{tab:transferability}
\setlength{\tabcolsep}{3.2pt}
\begin{tabular}{l|l|cc|cc|cc|cc}
\toprule
\multirow{2}{*}{Source}  & \multirow{2}{*}{Attack} & \multicolumn{2}{c}{ALBEF} & \multicolumn{2}{c}{TCL}    & \multicolumn{2}{c}{CLIP$_{ViT}$} & \multicolumn{2}{c}{CLIP$_{CNN}$} \\
\cmidrule(r){3-10}
                          &                         & TR   R@1    & IR   R@1    & TR   R@1    & IR   R@1     & TR   R@1     & IR   R@1     & TR   R@1     & IR   R@1     \\
\midrule
\multirow{5}{*}{ALBEF}   & SA    &   65.69    &  73.95   &   17.60    &         32.95   &  31.17  &    45.23       &    32.82    &     45.49    \\
                          & CA               &   77.16      &    83.86   &  15.21        &    29.49     &   23.60    &    36.48    &   25.12      &   38.89   \\
                          & SGA & 97.39(2.5) & 97.15(2.52) & 45.84(51.7) & 55.79(37.98) & 33.62(58.5) & 44.23(39.1)  &    36.27(53.5)     &    46.62(35.28)    \\
                          & \cellcolor{gray!30}FGA-T$_{aug}$    & \cellcolor{gray!30}\textbf{99.06(0.9)} & \cellcolor{gray!30}\textbf{99.02(0.9)}  & \cellcolor{gray!30}46.89(51.2) & \cellcolor{gray!30}58.02(35.7) & \cellcolor{gray!30}36.07(55.9)  & \cellcolor{gray!30}47.2(36.64)  &  \cellcolor{gray!30} 38.95(51.0)     &  \cellcolor{gray!30}   50.12(32.62)   \\
                          & \cellcolor{gray!30}MFGA-T$_{aug}$ & \cellcolor{gray!30}97.6(2.3) & \cellcolor{gray!30}98.15(1.64) & \cellcolor{gray!30}\textbf{52.27(45.9)} & \cellcolor{gray!30}\textbf{62.57(31.86)} & \cellcolor{gray!30}\textbf{36.93(55.4)} & \cellcolor{gray!30}\textbf{48.39(35.98)} & \cellcolor{gray!30}\textbf{39.72(50.1)} & \cellcolor{gray!30}\textbf{50.6(32.36)} \\

\midrule
\multirow{5}{*}{TCL}     & SA              &    20.13    &      36.48    &     84.72    &    86.07    &     31.29   &      44.65   & 33.33      &     45.80   \\
                          & CA               &    23.15    &       40.04   &   77.94    &   85.59     &     27.85    &       41.19   &    30.74    &      44.11 \\
                          & SGA    &    49.64(48.5)   &   59.85(34.78)  &   98.21(1.7)  &   98.79(1.1)   &   34.11(57.6)    &      44.68(38.64)  &     37.93(52.4)   &  48.47(34.02)\\
                        & \cellcolor{gray!30}FGA-T$_{aug}$ & \cellcolor{gray!30}44.84(53.4) & \cellcolor{gray!30}58.54(35.92) & \cellcolor{gray!30}\textbf{99.16(0.8)} & \cellcolor{gray!30}\textbf{99.21(0.68)} & \cellcolor{gray!30}35.71(56.5) & \cellcolor{gray!30}47.71(36.18) & \cellcolor{gray!30}39.59(51.0) & \cellcolor{gray!30}49.95(32.98) \\
                        & \cellcolor{gray!30}MFGA-T$_{aug}$ & \cellcolor{gray!30}\textbf{50.78(47.6)} & \cellcolor{gray!30}\textbf{63.05(32.26)} & \cellcolor{gray!30}98.31(1.6) & \cellcolor{gray!30}98.57(1.22) & \cellcolor{gray!30}\textbf{36.32(56.0)} & \cellcolor{gray!30}\textbf{48.94(35.36)} & \cellcolor{gray!30}\textbf{40.74(49.7)} & \cellcolor{gray!30}\textbf{50.5(32.66)} \\

\midrule
\multirow{5}{*}{CLIP$_{ViT}$} & SA              &    9.59   &    23.25    &    11.38   &      25.60    &     79.75     &    86.79     &     30.78   &     39.76       \\
                          &CA               &     10.57  &     24.33    &      11.94   &    26.69    &    93.25    &   95.86     &  32.52     &   41.82    \\
                          & SGA                     & 12.62(84.7) & 27.34(64.3) &   14.86(82.2)    &    29.83(60.64)   & 99.26(0.6)   & 99.0(0.64)   & 38.7(49.8)   & 47.51(32.32) \\
                          & \cellcolor{gray!30}FGA-T$_{aug}$                    & \cellcolor{gray!30}12.93(84.4) & \cellcolor{gray!30}28.84(62.7) &   \cellcolor{gray!30}14.12(82.7)    &   \cellcolor{gray!30}30.12(60.44)     & \cellcolor{gray!30}\textbf{99.39(0.5)}   & \cellcolor{gray!30}\textbf{99.74(0.18) } & \cellcolor{gray!30}42.78(47.3)  & \cellcolor{gray!30}48.68(31.82) \\
                          & \cellcolor{gray!30}MFGA-T$_{aug}$                   & \cellcolor{gray!30}\textbf{13.56(83.9)} & \cellcolor{gray!30}\textbf{30.05(61.7)} &  \cellcolor{gray!30}\textbf{14.96(81.8)}   &     \cellcolor{gray!30}\textbf{30.98(59.74)}   & \cellcolor{gray!30}99.26(0.6)   & \cellcolor{gray!30}99.52(0.36)  & \cellcolor{gray!30}\textbf{44.44(46.2)}  & \cellcolor{gray!30}\textbf{50.94(30.52)} \\
\midrule
\multirow{5}{*}{CLIP$_{CNN}$} & SA  & 8.55    &     23.41  &    12.64    &        26.12    &    28.34    &      39.43    &   91.44     &  95.44     \\
                          & CA   &    8.79    &      23.74  &    13.10    &        26.07    &    28.79    &      40.03  &  94.76    & 96.89   \\
                          & SGA  &   11.16(86.1) & 25.07(66.14) & 14.12(82.6) & 27.74(62.62) & 31.17(58.8) & 42.78(37.76) & 99.74(0.2) & 99.55(0.26) \\
                          & \cellcolor{gray!30}FGA-T$_{aug}$   & \cellcolor{gray!30}12.83(84.6) & \cellcolor{gray!30}26.29(64.9)  & \cellcolor{gray!30}14.23(82.9) & \cellcolor{gray!30}28.81(61.54) & \cellcolor{gray!30}35.34(55.5) & \cellcolor{gray!30}45.26(36.14) & \cellcolor{gray!30}\textbf{100.0(0.0)} & \cellcolor{gray!30}\textbf{99.93(0.04)} \\  
                          &\cellcolor{gray!30}MFGA-T$_{aug}$   &   \cellcolor{gray!30}\textbf{13.35(84.5)} & \cellcolor{gray!30}\textbf{27.48(63.88)} & \cellcolor{gray!30}\textbf{14.86(82.3)} & \cellcolor{gray!30}\textbf{30.1(60.52) } & \cellcolor{gray!30}\textbf{37.42(53.8)} & \cellcolor{gray!30}\textbf{47.2(35.04)}  & \cellcolor{gray!30}100.0(0.0) & \cellcolor{gray!30}99.90(0.06)        \\
\bottomrule
\end{tabular}
\end{table*}

The experimental results are shown in Table~\ref{tab:transferability}. 
We observe the following phenomena:
(1) SA, CA, and SGA attack the visual modality based on feature deviation. 
SGA designs a more advanced set-level feature deviation and introduces data augmentation, improving both white-box and black-box attack effects on the baseline.
(2) FGA-T\(_{aug}\) based on feature guidance, improves SGA further, simultaneously enhancing both white-box and black-box attack effects again.  
(3) MFGA-T\(_{aug}\) slightly reduces the white-box attack effect but further improves adversarial transferability, which is consistent with the observations in \cite{mi}.
(4) Attacks based on ALBEF transfer better to TCL than to CLIP because ALBEF and TCL only have differences in parameters, while ALBEF and CLIP are completely different models. 
The same logic applies to attacks based on TCL.
(5) Attacks based on CLIP\(_{ViT}\) transfer better to CLIP\(_{CNN}\) than to ALBEF or TCL because the model difference between CLIP\(_{ViT}\) and CLIP\(_{CNN}\) is obviously smaller than the difference with ALBEF or TCL. 
The same logic applies to attacks based on CLIP\(_{CNN}\).  

\begin{figure}[t]
  \centering
  \includegraphics[width=\linewidth]{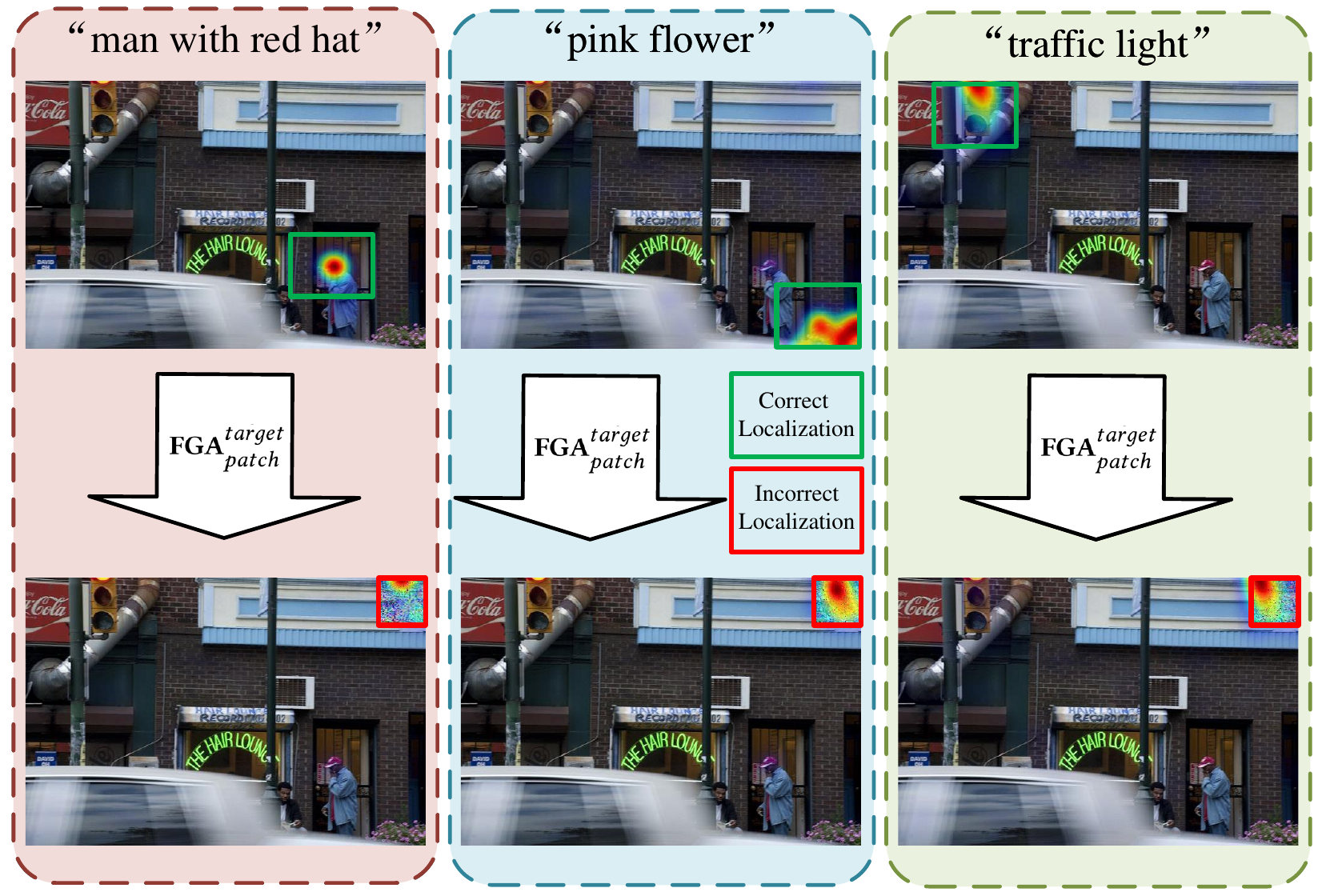}
  \caption{Before the attack, ALBEF can accurately localize image content based on textual cues. After FGA\(_{patch}^{target}\), ALBEF's attention is always erroneously focused on the patch.}
  \label{tab:targeted}
  \Description{Before the attack, ALBEF can accurately localize image content based on textual cues. After FGA\(_{patch}^{target}\), ALBEF's attention is always erroneously focused on the patch.}
\end{figure}

\begin{figure}[t]
  \centering
  \includegraphics[width=\linewidth]{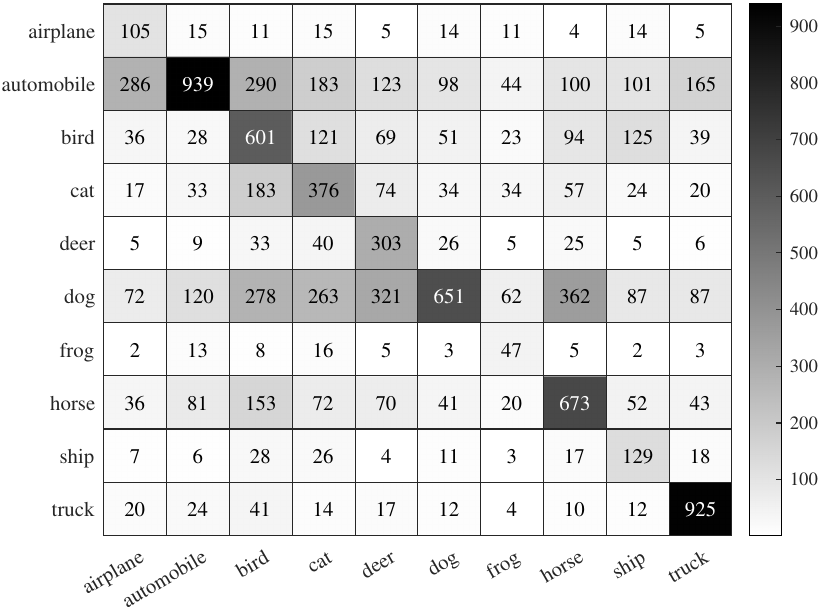}
  \caption{Each row represents the predicted category for \(v\) excluding the correct category \(y\), and each column represents the predicted category for \(v'\). }
  \label{tab:heatmap}
  \Description{Each row represents the predicted category for \(v\) excluding the correct category \(y\), and each column represents the predicted category for \(v'\). }
\end{figure}

\subsection{Visualization of Targeted Patch FGA}
FGA pushes \(E_v(v')\) away from matching text embeddings. 
Conversely, we can also push \(E_v(v')\) closer to a specified text embedding to produce a predetermined error. 
In unimodal scenarios, this form of attack is called the targeted attack. 
For example, we have a text set \(\{t\}_{i=1}^n\) and want to push \(E(v')\) closer to a specified text \(t_k\). 
In this case, we need to maximize the following function:
\begin{equation}
  L_{gui}^{target} = ln\left( \frac{\exp(E(v') \cdot E_t(t_k))}{\sum_{i=1}^{n}\exp(E(v') \cdot E_t(t_i))} \right) 
\end{equation}

We add perturbation to the clean image \(v\) in patch form, maximizing \(L_{gui}^{target}\) to obtain the adversarial patch image \(v'\). 
We execute FGA\(_{patch}^{target}\) on the ALBEF model and compute Grad-CAM visualizations on the self-attention maps. 
As shown in Figure~\ref{tab:targeted}, by guiding \(E(v')\) closer to the prompt text through FGA\(_{patch}^{target}\), ALBEF's attention area for \(v'\) is concentrated on the patch, resulting in a misunderstanding.

\subsection{FGA's Principle of Proximity}
In subection~\ref{sec:Feature Guidance}, it is mentioned that \(\frac{\partial L_{gui}}{\partial E(v')}\) not only ``guides \(E(v')\) away from \(\{\omega_{y_i}\}_{i=1}^n\)'', but also ``selects a nearby guiding vector that does not belong to the set \(\{\omega_{y_i}\}_{i=1}^n\) and moves closer to it''. 
The performance decline of VLP models on various V+L downstream tasks in previous experiments sufficiently demonstrates the former.
We further prove the latter based on the ZC task and the CLIP model.
In the ZC task, we collect the text set \(T = \{t_i\}_{i=1}^c\) and use it to construct the guiding vectors \(\{E_t(t_i)\}_{i=1}^c\). 
FGA encourages \(E_v(v')\) to move away from \(E_t(t_y)\), where \(y\) is the true category, and simultaneously encourages \(E_v(v')\) to move closer to the nearest vector from \(\{E_t(t_i)\}_{i=1, i \neq y}^c\), meaning that in an ideal situation:
\begin{equation}
  \underset{i,i\neq y}{argmax} \frac{E_v(v)\cdot E_t(t_i)}{\left \| E_v(v)  \right \| \left \| E_t(t_i) \right \| } = \underset{i}{argmax} \frac{E_v(v')\cdot E_t(t_i)}{\left \| E_v(v')  \right \| \left \| E_t(t_i) \right \| }
\end{equation}
In simpler terms, the category predicted for the clean image \(v\), excluding the true category \(y\), will be the category predicted for the adversarial image \(v'\).
Based on the CIFAR-10 dataset, we present the statistical results in Figure~\ref{tab:heatmap}. 
In an ideal situation, all positions except the main diagonal should be zero. 
We observe that the actual situation is close to the ideal. 
This indicates that the FGA attack indeed tends to guide ``\(E(v')\) to move closer to the nearest vector from \(\{E_t(t_i)\}_{i=1, i \neq y}^c\)''.
In fact, this principle of proximity promotes \(v'\) to automatically choose the nearest decision boundary to cross, which is also one of the reasons for the success of FGA.

\begin{figure}[t]
  \centering
  \includegraphics[width=\linewidth]{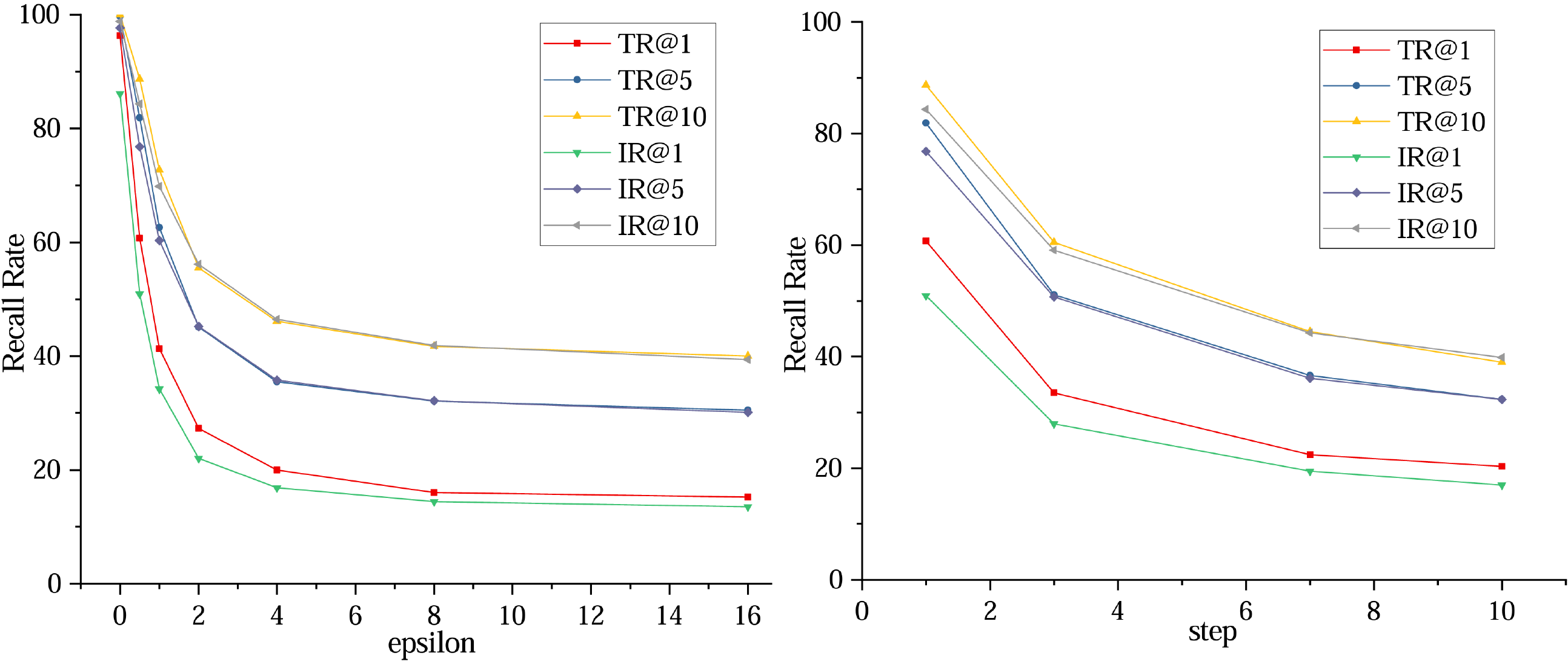}
  \caption{On the left, we fix \(step\) at 1 and investigate the impact of \(\epsilon\). On the right, we fix \(\epsilon\) at 0.5 and explore the effect of \(step\).}
  \label{tab:ablation}
  \Description{Each row represents the predicted category for \(v\) excluding the correct category \(y\), and each column represents the predicted category for \(v'\). }
\end{figure}

\subsection{Ablation Experiments}

We investigate the impact of the number of iterations (\(step\)) and intensity of noise (\(\epsilon\)) based on the image-text retrieval task, Flickr30k dataset and the BEiT-3 model, as shown in Fig~\ref{tab:ablation}. 
We can observe that as \(\epsilon\) and \(step\) increase, the effectiveness of the attack gradually strengthens and tends to converge. 
Specific experimental configurations are detailed in Appendix C.

\section{CONCLUSION}
In this paper, we attempt to construct a unified understanding of adversarial vulnerability regarding unimodal models and VLP models. 
We abstract visual modality attack into a feature guidance form and combine it with text attack and other enhancement mechanisms to establish a general baseline for exploring the security of the VLP domain.
In fact, our approach is theoretically orthogonal to many other attack schemes in the unimodal domain, which facilitates further exploration of the vulnerabilities of VLP models and the design of defence algorithms in subsequent work.
We hope our code can be beneficial to the community.

\balance
\printbibliography

@String{Computing = "Computing" }

@String{Computer = "{IEEE} Computer" }

@inproceedings{li1,
author = {Li, Wenrui and Ma, Zhengyu and Deng, Liang-Jian and Wang, Penghong and Shi, Jinqiao and Fan, Xiaopeng},
title = {Reservoir Computing Transformer for Image-Text Retrieval},
year = {2023}, 
booktitle={Proceedings of the 31st ACM International Conference on Multimedia (ACM MM)},
pages={5605--5613}
}

@inproceedings{li2,
author = {Li, Wenrui and Zhao, Xi-Le and Ma, Zhengyu and Wang, Xingtao and Fan, Xiaopeng and Tian, Yonghong},
title = {Motion-Decoupled Spiking Transformer for Audio-Visual Zero-Shot Learning},
booktitle={Proceedings of the 31st ACM International Conference on Multimedia (ACM MM)},
pages={3994--4002},
year = {2023},}

@inproceedings{vit,
  title={An Image is Worth 16x16 Words: Transformers for Image Recognition at Scale},
  author={Dosovitskiy, Alexey and Beyer, Lucas and Kolesnikov, Alexander and Weissenborn, Dirk and Zhai, Xiaohua and Unterthiner, Thomas and Dehghani, Mostafa and Minderer, Matthias and Heigold, Georg and Gelly, Sylvain and others},
  booktitle={International Conference on Learning Representations (ICLR)},
  year = {2023},
}

@inproceedings{albef,
      title={Align before Fuse: Vision and Language Representation Learning with Momentum Distillation}, 
      author={Li, Junnan and Selvaraju, Ramprasaath and Gotmare, Akhilesh and Joty, Shafiq and Xiong, Caiming and Hoi, Steven Chu Hong},
      booktitle={Proceedings of the 35th
 International Conference on Neural Information Processing Systems (NeurIPS)},
pages={9694--9705},
year = {2021},
}

@InProceedings{tcl,
    author    = {Yang, Jinyu and Duan, Jiali and Tran, Son and Xu, Yi and Chanda, Sampath and Chen, Liqun and Zeng, Belinda and Chilimbi, Trishul and Huang, Junzhou},
    title     = {Vision-Language Pre-Training With Triple Contrastive Learning},
    booktitle = {Proceedings of the IEEE/CVF Conference on Computer Vision and Pattern Recognition (CVPR)},
pages={15671--15680},
year = {2022},
}

@inproceedings{beit3,
title={Image as a Foreign Language: {BEiT} Pretraining for Vision and Vision-Language Tasks},
author={Wang, Wenhui and Bao, Hangbo and Dong, Li and Bjorck, Johan and Peng, Zhiliang and Liu, Qiang and Aggarwal, Kriti and Mohammed, Owais Khan and Singhal, Saksham and Som, Subhojit and others},
booktitle={Proceedings of the IEEE/CVF Conference on Computer Vision and Pattern Recognition (CVPR)},
pages={19175--19186},
year={2023},
}

@inproceedings{beit2,
title={{BEiT v2}: Masked Image Modeling with Vector-Quantized Visual Tokenizers},
author={Zhiliang Peng and Li Dong and Hangbo Bao and Qixiang Ye and Furu Wei},
booktitle={arXiv preprint arXiv:2208.06366},
year = {2022},
}

@inproceedings{beit,
title={{BEiT}: {BERT} Pre-Training of Image Transformers},
author={Bao, Hangbo and Dong, Li and Piao, Songhao and Wei, Furu},
booktitle={International Conference on Learning Representations (ICLR)},
year={2022},
}

@InProceedings{vilt,
  title = 	 {ViLT: Vision-and-Language Transformer Without Convolution or Region Supervision},
  author =       {Kim, Wonjae and Son, Bokyung and Kim, Ildoo},
  booktitle = 	 {Proceedings of the International Conference on Machine
 Learning (ICML)},
pages={5583--5594},
  year = 	 {2021},
}

@inproceedings{vlmo,
 author = {Bao, Hangbo and Wang, Wenhui and Dong, Li and Liu, Qiang and Mohammed, Owais Khan and Aggarwal, Kriti and Som, Subhojit  and Piao, Songhao and Wei, Furu},
 booktitle = {Proceedings of the 36th
 International Conference on Neural Information Processing Systems (NeurIPS)},
 year = {2022},
pages={32897--32912},
title = {VLMo: Unified Vision-Language Pre-Training with Mixture-of-Modality-Experts},
}

@InProceedings{clip,
  title = 	 {Learning Transferable Visual Models From Natural Language Supervision},
  author =       {Radford, Alec and Kim, Jong Wook and Hallacy, Chris and Ramesh, Aditya and Goh, Gabriel and Agarwal, Sandhini and Sastry, Girish and Askell, Amanda and Mishkin, Pamela and Clark, Jack and Krueger, Gretchen and Sutskever, Ilya},
  booktitle = 	 {Proceedings of the International Conference on Machine
 Learning (ICML)},
pages={8748--8763},
  year = 	 {2021},
}

@inproceedings{fgsm,
title	= {Explaining and Harnessing Adversarial Examples},
author	= {Goodfellow, Ian J and Shlens, Jonathon and Szegedy, Christian},
year	= {2015},
booktitle	= {International Conference on Learning Representations (ICLR)},
}

@inproceedings{pgd,
title={Towards Deep Learning Models Resistant to Adversarial Attacks},
author={Madry, Aleksander and Makelov, Aleksandar and Schmidt, Ludwig and Tsipras, Dimitris and Vladu, Adrian},
booktitle={International Conference on Learning Representations (ICLR)},
year={2018},
}

@INPROCEEDINGS{cw,
  author={Carlini, Nicholas and Wagner, David},
  booktitle={IEEE Symposium on Security and Privacy (S\&P)}, 
  title={Towards Evaluating the Robustness of Neural Networks}, 
pages={39--57},
  year={2017},
}

@inproceedings{apgd1,
    title={Mind the box: $l_1$-APGD for sparse adversarial attacks on image classifiers}, 
    author={Croce, Francesco and Hein, Matthias},
    booktitle={Proceedings of the International Conference on Machine
 Learning (ICML)},
pages={2201--2211},
    year={2021}
}

@INPROCEEDINGS{deepfool,
  author={Moosavi-Dezfooli, Seyed-Mohsen and Fawzi, Alhussein and Frossard, Pascal},
  booktitle={Proceedings of the IEEE Conference on Computer Vision and Pattern Recognition (CVPR)}, 
  title={DeepFool: A Simple and Accurate Method to Fool Deep Neural Networks}, 
pages={2574--2582},
  year={2016},
}

@inproceedings{bertatk,
    title = {BERT-ATTACK: Adversarial Attack Against BERT Using BERT},
    author = {Li, Linyang  and
      Ma, Ruotian  and
      Guo, Qipeng  and
      Xue, Xiangyang  and
      Qiu, Xipeng},
    booktitle = {Proceedings of the 2020 Conference on Empirical Methods in Natural Language Processing (EMNLP)},
pages = "6193--6202",
    year = {2020},

}

@INPROCEEDINGS{mi,
  author={Dong, Yinpeng and Liao, Fangzhou and Pang, Tianyu and Su, Hang and Zhu, Jun and Hu, Xiaolin and Li, Jianguo},
  booktitle={Proceedings of the IEEE Conference on Computer Vision and Pattern Recognition (CVPR)},
  pages={9185--9193},
  title={Boosting Adversarial Attacks with Momentum}, 
  year={2018}}

@INPROCEEDINGS{di,
  author={Xie, Cihang and Zhang, Zhishuai and Zhou, Yuyin and Bai, Song and Wang, Jianyu and Ren, Zhou and Yuille, Alan L.},
  booktitle={Proceedings of the IEEE/CVF Conference on Computer Vision and Pattern Recognition (CVPR)}, 
  pages={2730--2739},
  title={Improving Transferability of Adversarial Examples With Input Diversity}, 
  year={2019},}

@INPROCEEDINGS{ti,
  author={Dong, Yinpeng and Pang, Tianyu and Su, Hang and Zhu, Jun},
  booktitle={Proceedings of the IEEE/CVF Conference on Computer Vision and Pattern Recognition (CVPR)}, 
  title={Evading Defenses to Transferable Adversarial Examples by Translation-Invariant Attacks}, 
pages={4312--4321},
  year={2019},}

@inproceedings{ni,
title={Nesterov Accelerated Gradient and Scale Invariance for Adversarial Attacks},
author={Lin, Jiadong and Song, Chuanbiao and He, Kun and Wang, Liwei and Hopcroft, John E},
booktitle={International Conference on Learning Representations (ICLR)},
year={2020},}

@INPROCEEDINGS{vmi,
  author={Wang, Xiaosen and He, Kun},
  booktitle={Proceedings of the IEEE/CVF Conference on Computer Vision and Pattern Recognition (CVPR)}, 
  title={Enhancing the Transferability of Adversarial Attacks through Variance Tuning}, 
pages={1924--1933},
  year={2021},}

@inproceedings{coattack,
author = {Zhang, Jiaming and Yi, Qi and Sang, Jitao},
title = {Towards Adversarial Attack on Vision-Language Pre-training Models},
year = {2022},
  booktitle={Proceedings of the 30th ACM International Conference on Multimedia (ACM MM)},
  pages={5005--5013},
}

@inproceedings{advclip,
author = {Zhou, Ziqi and Hu, Shengshan and Li, Minghui and Zhang, Hangtao and Zhang, Yechao and Jin, Hai},
title = {Adv{CLIP}: Downstream-agnostic Adversarial Examples in Multimodal Contrastive Learning},
  booktitle={Proceedings of the 31st ACM International Conference on Multimedia (ACM MM)},
  pages={6311--6320},
year = {2023},
}

@InProceedings{sga,
    author    = {Lu, Dong and Wang, Zhiqiang and Wang, Teng and Guan, Weili and Gao, Hongchang and Zheng, Feng},
    title     = {Set-level Guidance Attack: Boosting Adversarial Transferability of Vision-Language Pre-training Models},
  booktitle={Proceedings of the IEEE/CVF International Conference on Computer Vision (ICCV)},
  pages={102--111},
    year      = {2023},
}

@INPROCEEDINGS{targetedatk,
  author={Li, Maosen and Deng, Cheng and Li, Tengjiao and Yan, Junchi and Gao, Xinbo and Huang, Heng},
  booktitle={Proceedings of the IEEE/CVF Conference on Computer Vision and Pattern Recognition (CVPR)},  
  title={Towards Transferable Targeted Attack}, 
pages={641--649},
  year={2020},}

@inproceedings{blip,
      title={BLIP: Bootstrapping Language-Image Pre-training for Unified Vision-Language Understanding and Generation}, 
      author={Li, Junnan and Li, Dongxu and Xiong, Caiming and Hoi, Steven},
      year={2022},
pages={12888--12900},
  booktitle = 	 {Proceedings of the International Conference on Machine
 Learning (ICML)},
}

@InProceedings{lavan,
  title = 	 {La{VAN}: Localized and Visible Adversarial Noise},
  author =       {Karmon, Danny and Zoran, Daniel and Goldberg, Yoav},
  booktitle = 	 {Proceedings of the International Conference on Machine
 Learning (ICML)},
  year = 	 {2018},
pages={2507--2515},
}

@INPROCEEDINGS{moco,
  author={He, Kaiming and Fan, Haoqi and Wu, Yuxin and Xie, Saining and Girshick, Ross},
  booktitle={Proceedings of the IEEE/CVF Conference on Computer Vision and Pattern Recognition (CVPR)}, 
  title={Momentum Contrast for Unsupervised Visual Representation Learning}, 
pages={9729--9738},
  year={2020},
}

@inproceedings{simclr,
author = {Chen, Ting and Kornblith, Simon and Norouzi, Mohammad and Hinton, Geoffrey},
title = {A simple framework for contrastive learning of visual representations},
year = {2020},
  booktitle = 	 {Proceedings of the International Conference on Machine
 Learning (ICML)},
pages={1597--1607},
}

@inproceedings{byol,
author = {Grill, Jean-Bastien and Strub, Florian and Altch\'{e}, Florent and Tallec, Corentin and Richemond, Pierre H. and Buchatskaya, Elena and Doersch, Carl and Pires, Bernardo Avila and Guo, Zhaohan Daniel and Azar, Mohammad Gheshlaghi and Piot, Bilal and Kavukcuoglu, Koray and Munos, R\'{e}mi and Valko, Michal},
title = {Bootstrap your own latent a new approach to self-supervised learning},
year = {2020},
booktitle = {Proceedings of the 34th
 International Conference on Neural Information Processing Systems (NeurIPS)},
pages={21271--21284},
}

@article{cifar10,
  title={Learning multiple layers of features from tiny images},
  author={ Krizhevsky, A.  and  Hinton, G. },
  journal={Handbook of Systemic Autoimmune Diseases},
  year={2009},
}

@INPROCEEDINGS{snlive,
      title={{Visual Entailment}: A Novel Task for Fine-Grained Image Understanding}, 
      author={Ning Xie and Farley Lai and Derek Doran and Asim Kadav},
      year={2019},
booktitle={arXiv preprint:1901.06706}
}

@INPROCEEDINGS{vqa1,
  author={Goyal, Yash and Khot, Tejas and Summers-Stay, Douglas and Batra, Dhruv and Parikh, Devi},
  booktitle={Proceedings of the IEEE Conference on Computer Vision and Pattern Recognition (CVPR)}, 
  title={Making the V in VQA Matter: Elevating the Role of Image Understanding in Visual Question Answering}, 
pages={6904--6913},
  year={2017},}

@InProceedings{refcoco,
author={Yu, Licheng
and Poirson, Patrick
and Yang, Shan
and Berg, Alexander C.
and Berg, Tamara L.},
title={Modeling Context in Referring Expressions},
pages={69--85},
booktitle={Proceedings of the 14th European Conference on Computer Vision (ECCV)},
year={2016},
}

@inproceedings{depatch,
  title={Shape matters: deformable patch attack},
  author={Chen, Zhaoyu and Li, Bo and Wu, Shuang and Xu, Jianghe and Ding, Shouhong and Zhang, Wenqiang},
  booktitle={Proceedings of the 17th European Conference on Computer Vision (ECCV)},
  pages={529--548},
  year={2022},
}

@inproceedings{grad-cam,
  title={Grad-cam: Visual Explanations from Deep Networks via Gradient-based Localization},
  author={Selvaraju, Ramprasaath R and Cogswell, Michael and Das, Abhishek and Vedantam, Ramakrishna and Parikh, Devi and Batra, Dhruv},
  booktitle={Proceedings of the IEEE International Conference on Computer Vision (ICCV)},
  pages={618--626},
  year={2017}
}

@inproceedings{nlvr2,
    title = {A Corpus for Reasoning about Natural Language Grounded in Photographs},
    author = {Suhr, Alane  and
      Zhou, Stephanie  and
      Zhang, Ally  and
      Zhang, Iris  and
      Bai, Huajun  and
      Artzi, Yoav},
    booktitle = {Proceedings of the 57th Annual Meeting of the Association for Computational Linguistics (ACL)},
pages = "6418--6428",
    year = {2019},
}

@INPROCEEDINGS{imagenet,
  author={Deng, Jia and Dong, Wei and Socher, Richard and Li, Li-Jia and Kai, Li and Li, Fei-Fei},
    booktitle={Proceedings of the IEEE Conference on Computer Vision and Pattern Recognition (CVPR)}, 
  title={{ImageNet}: A large-scale hierarchical image database}, 
pages={248--255},
  year={2009},}

@INPROCEEDINGS{flickr30k,
  author={Plummer, Bryan A. and Wang, Liwei and Cervantes, Chris M. and Caicedo, Juan C. and Hockenmaier, Julia and Lazebnik, Svetlana},
  booktitle={Proceedings of the IEEE International Conference on Computer Vision (ICCV)}, 
  title={Flickr30k Entities: Collecting Region-to-Phrase Correspondences for Richer Image-to-Sentence Models}, 
  pages={2641--2649},
  year={2015},}

@InProceedings{coco,
author={Lin, Tsung-Yi
and Maire, Michael
and Belongie, Serge
and Hays, James
and Perona, Pietro
and Ramanan, Deva
and Doll{\'a}r, Piotr
and Zitnick, C. Lawrence},
title={Microsoft {COCO}: Common Objects in Context},
booktitle={Proceedings of the 13th European Conference on Computer Vision (ECCV)},
pages={740--755},
year={2014},
}

@article{vqa2,
  title={Mra-net: Improving vqa via multi-modal relation attention network},
  author={Peng, Liang and Yang, Yang and Wang, Zheng and Huang, Zi and Shen, Heng Tao},
  journal={IEEE Transactions on Pattern Analysis and Machine Intelligence (TPAMI)},
  pages={318--329},
  year={2020},
}

@inproceedings{vln1,
  title={Vision-and-language navigation: Interpreting visually-grounded navigation instructions in real environments},
  author={Anderson, Peter and Wu, Qi and Teney, Damien and Bruce, Jake and Johnson, Mark and S{\"u}nderhauf, Niko and Reid, Ian and Gould, Stephen and Van Den Hengel, Anton},
  booktitle={Proceedings of the IEEE Conference on Computer Vision and Pattern Recognition (CVPR)},
  pages={3674--3683},
  year={2018}
}

@inproceedings{vln2,
  title={Reinforced cross-modal matching and self-supervised imitation learning for vision-language navigation},
  author={Wang, Xin and Huang, Qiuyuan and Celikyilmaz, Asli and Gao, Jianfeng and Shen, Dinghan and Wang, Yuan-Fang and Wang, William Yang and Zhang, Lei},
  booktitle={Proceedings of the IEEE/CVF Conference on Computer Vision and Pattern Recognition (CVPR)},
  pages={6629--6638},
  year={2019}
}

@inproceedings{arra,
  title={Arra: Absolute-relative ranking attack against image retrieval},
  author={Li, Siyuan and Xu, Xing and Zhou, Zailei and Yang, Yang and Wang, Guoqing and Shen, Heng Tao},
  booktitle={Proceedings of the 30th ACM International Conference on Multimedia (ACM MM)},
  pages={610--618},
  year={2022}
}

@article{itr1,
  title={Adaptive semi-supervised feature selection for cross-modal retrieval},
  author={Yu, En and Sun, Jiande and Li, Jing and Chang, Xiaojun and Han, Xian-Hua and Hauptmann, Alexander G},
  journal={IEEE Transactions on Multimedia (TMM)},
  pages={1276--1288},
  year={2018},
}

@article{itr2,
  title={Upgrading the newsroom: An automated image selection system for news articles},
  author={Liu, Fangyu and Lebret, R{\'e}mi and Orel, Didier and Sordet, Philippe and Aberer, Karl},
  journal={ACM Transactions on Multimedia Computing, Communications, and Applications (TOMM)},
  pages={1--28},
  year={2020},
}

@article{itr3,
  title={Less is better: Exponential loss for cross-modal matching},
  author={Wei, Jiwei and Yang, Yang and Xu, Xing and Song, Jingkuan and Wang, Guoqing and Shen, Heng Tao},
  journal={IEEE Transactions on Circuits and Systems for Video Technology (TCSVT)},
  year={2023},
  pages={5271-5280},
}

@inproceedings{msd,
author = {Maini, Pratyush and Wong, Eric and Kolter, J. Zico},
title = {Adversarial robustness against the union of multiple perturbation models},
year = {2020},
booktitle = {Proceedings of the International Conference on Machine Learning (ICML)},
pages={6640-6650},
}

@inproceedings{eat,
  title={Adversarial Robustness against Multiple and Single $l_p$-Threat Models via Quick Fine-Tuning of Robust Classifiers},
  author={Croce, Francesco and Hein, Matthias},
booktitle = {Proceedings of the International Conference on Machine Learning (ICML)},
  pages={4436--4454},
  year={2022},
}

@inproceedings{ncat,
  title={Toward Efficient Robust Training against Union of $\ell_p $ Threat Models},
  author={Sriramanan, Gaurang and Gor, Maharshi and Feizi, Soheil},
 booktitle = {Proceedings of the 36th
 International Conference on Neural Information Processing Systems (NeurIPS)},
  pages={25870--25882},
  year={2022}
}

@article{max,
  title={Adversarial training and robustness for multiple perturbations},
  author={Tramer, Florian and Boneh, Dan},
  journal={Advances in neural information processing systems},
  volume={32},
  year={2019}
}

@INPROCEEDINGS{rq1,
  author={Chen, Zhuangzhuang and Zhang, Jin and Lai, Zhuonan and Chen, Jie and Liu, Zun and Li, Jianqiang},
    booktitle = {Proceedings of the IEEE/CVF Conference on Computer Vision and Pattern Recognition (CVPR)},
  title={Geometry-Aware Guided Loss for Deep Crack Recognition}, 
  year={2022},
  pages={4693-4702},}

@INPROCEEDINGS {rq2,
author = {Z. Chen and J. Zhang and Z. Lai and G. Zhu and Z. Liu and J. Chen and J. Li},
booktitle = {IEEE/CVF International Conference on Computer Vision (ICCV)},
title = {The Devil is in the Crack Orientation: A New Perspective for Crack Detection},
year = {2023},
pages = {6630-6640},
}

@ARTICLE{lwr3,
  author={Li, Wenrui and Wang, Penghong and Xiong, Ruiqin and Fan, Xiaopeng},
  journal={IEEE Transactions on Image Processing}, 
  title={Spiking Tucker Fusion Transformer for Audio-Visual Zero-Shot Learning}, 
  year={2024},
  pages={1-1},}
\newpage
\appendix

\section{Attack Details} 
\label{sec:Attack Details}
Three key elements are required to implement FGA, namely an image encoder, guiding vectors, and guiding labels. Below, we will elaborate on the construction of these elements in four scenarios: VE, VQA, VG, and VR.
\subsection{Visual Entailment}

\paragraph{\textbf{Task Detail}} We conduct the attack experiment on the VE task \cite{snlive} with ALBEF \cite{albef} which treats VE as a three-classification problem and connects a multi-layer perceptron (MLP) after the [CLS] vector. 
The input layer and all hidden layers of the MLP constitute \(P\), obtaining the image encoder \(E(v|t) = P(E_m(E_v(v), E_t(t)))\).
The output layer of the MLP is a linear layer, whose weight matrix is \(W = \{\omega_0, \omega_1, \omega_2\}\).
The three guiding vectors are associated with three categories ``contradiction, neutral and entailment'', and the label \(y \in \{0,1,2\}\) of the input image-text pair \((v,t)\) provides the direction of the attack, that is, guiding \(E(v'|t)\) the embedding of adversarial image \(v'\) deviates from the guiding vector \(\omega_y\).

\paragraph{\textbf{Dataset Detail}} The SNLI-VE dataset \cite{snlive} is a benchmark for visual entailment, which aims to determine whether an image supports, contradicts, or is neutral to a given natural language statement. 
This task extends the concept of natural language inference (NLI) to the visual domain, presenting challenges in image and text understanding. 
The dataset is constructed based on two existing datasets: SNLI (Stanford Natural Language Inference) and Flick30k. We use its test split, which contains 1000 images, 5973 entailment texts, 5964 neutral texts, and 5964 contradiction texts.

\subsection{Visual Question Answering}
\paragraph{\textbf{Task Detail}} We conduct the attack experiment on the VQA task \cite{vqa1} with ALBEF which performs this task in the manner of text generation. 
The image-question pair is fed into ALBEF to extract the fused embedding, which is then sent to a decoder to generate an answer. 
The dictionary size of the decoder is 30522, so the end of the decoder is a linear classification head, with weight matrix \(\{\omega_i\}_{i=0}^{30521}\). 
The VQA 2.0 dataset \cite{vqa1} provides 3,128 candidate answers. 
To align with this task, ALBEF only considers 3,128 output possibilities. 
We follow this by selecting 3,128 vectors from the weights of the linear layer to form the guiding vectors \(\{\omega_i\}_{i=0}^{3127}\), each corresponding to an answer. 
To perform FGA, we denote the decoder excluding the linear classification head as \(P\), and we still lack guiding labels. 
For convenience, we directly use the network's prediction results as the guiding labels, which is \(argmax_i (P(E_m(E_v(v'), E_t(t)))\cdot \omega_i)\).

\paragraph{\textbf{Dataset Detail}} The VQA2.0 dataset includes images from the MS COCO (Microsoft Common Objects in Context) dataset \cite{coco}, providing a diverse set of real-world images depicting various objects, scenes, and activities. For each image, multiple questions are generated, covering a wide range of topics such as object recognition, counting, colour identification, spatial relationships, and more. Each question is accompanied by multiple answers, provided by different human annotators. The answers can be in the form of single words, phrases, or numbers. It contains 83k images for training, 41k for validation, and 81k for test. We conduct attack tests based on the test-dev and test-std splits.

\subsection{Visual Grounding}
\paragraph{\textbf{Task Detail}} We conduct the attack experiment on the VE task with ALBEF which extends Grad-CAM \cite{grad-cam} to acquire heatmaps and use them to rank the detected proposals provided in advance. 
During this task, after the fused encoder, ALBEF is followed by a linear image-text matching binary classifier, the weight matrix of which is \(W = \{\omega_0, \omega_1\}\). 
The larger the inner product between the fused embedding and \(\omega_1\), the more the input image-text pair \((v,t)\) matches. 
ALBEF backpropagates the gradient based on the loss value \(\text{Em}(E_v(v), E_t(t)) \cdot \omega_1\), obtains the heatmap, and then performs the VG task. 
Consequently, we use FGA to guide \(\text{Em}(E_v(v'), E_t(t))\) away from \(\omega_1\) as the attack strategy.

\paragraph{\textbf{Dataset Detail}} RefCOCO+ \cite{refcoco} is a dataset designed for referring expression comprehension in the context of images. 
It is an extension of the original RefCOCO dataset and specifically aims at addressing the challenge of grounding referring expressions that require fine-grained distinctions between objects.
The key components of the RefCOCO+ dataset are:
(1) \textbf{Images}: The dataset uses images from the Microsoft COCO (Common Objects in Context) dataset, which contains a wide variety of everyday scenes with multiple objects. (2) \textbf{Referring Expressions}: For each image, there are several referring expressions provided by human annotators. These expressions describe specific objects or groups of objects in the image. (3) \textbf{Object Annotations}: Each referring expression is associated with an object annotation, a bounding box that identifies the location of the referred object in the image.

\subsection{Visual Grounding}
\paragraph{\textbf{Task Detail}}
We perform the VR task based on the BEiT3 model \cite{beit3}. 
In this task, the input example pair of the model is \((v_0, v_1, t, y)\), where \(y \in \{0, 1\}\). 
\(y = 1\) means that the text matches at least one of two images.
BEiT3 splits an example pair into two image-text pairs \((v_0, t)\) and \((v_1, t)\) as inputs, thereby extracting two fused embeddings. 
After concatenating the two embeddings and performing operations such as nonlinear projection, the final feature vector is obtained. 
This feature vector is fed into a binary classifier, whose weight matrix is \(\{\omega_0, \omega_1\}\). 
At this point, we only need to guide the feature vector away from the guiding vector \(\omega_y\) through FGA, and simultaneously update the input images \(v_0, v_1\) along the gradient direction to obtain the adversarial images \(v_0'\) and \(v_1'\).

\paragraph{\textbf{Dataset Detail}} 
NLVR2 \cite{nlvr2} (Natural Language for Visual Reasoning for Real) is a natural language processing dataset designed for the visual reasoning task. 
It aims to evaluate models' ability to reason about visual information combined with natural language descriptions. 
NLVR2 is an extended version of the NLVR dataset, featuring more images and more complex language descriptions. 
The NLVR2 dataset contains approximately 107,000 human-written sentences describing visual relationships in a set of images. 
Each sample includes a sentence and a pair of images. 
The content described in the sentence may match one of the images, both, or neither. 
The task for models is to determine whether the sentence correctly describes at least one of the images.
This dataset is used for various vision-language tasks, such as visual question answering, image-text matching, and multimodal reasoning. 
NLVR2 advances the research and development of vision-language models' reasoning capabilities by providing more challenging samples and complex language descriptions.

\begin{table}[t]
\caption{
The experimental results when \(step = 1\). The reported values are recall rates. Lower is better.
}
\label{tab:step1}
\begin{tabular}{c|cccccc}
\toprule
\multirow{2}{*}{\(\epsilon\ (\ell_\infty)\)}  &\multicolumn{3}{c}{Test Retrieval} & \multicolumn{3}{c}{Image Retrieval} \\
\cmidrule(r){2-4} \cmidrule(r){5-7}
\multirow{2}{*}{}        & R@1   & R@5   & R@10  & R@1   & R@5   & R@10    \\
\midrule

w/o atk & 96.30 & 99.70 & 100.00 & 86.14 & 97.68 & 98.82 \\
0.5     & 60.70 & 81.90 & 88.70  & 50.94 & 76.80 & 84.32 \\
1       & 41.30 & 62.60 & 72.70  & 34.24 & 60.32 & 69.86 \\
2       & 27.30 & 45.10 & 55.50  & 22.04 & 45.22 & 56.14 \\
4       & 20.00 & 35.50 & 46.10  & 16.88 & 35.74 & 46.46 \\
8       & 16.00 & 32.10 & 41.70  & 14.42 & 32.14 & 41.86 \\
16      & 15.20 & 30.50 & 40.00  & 13.54 & 30.08 & 39.40 \\

\bottomrule
\end{tabular}
\end{table}

\begin{table}[t]
\caption{
The experimental results when \(step = 3\). The reported values are recall rates. Lower is better.
}
\label{tab:step3}
\begin{tabular}{c|cccccc}
\toprule
\multirow{2}{*}{\(\epsilon\ (\ell_\infty)\)}  &\multicolumn{3}{c}{Test Retrieval} & \multicolumn{3}{c}{Image Retrieval} \\
\cmidrule(r){2-4} \cmidrule(r){5-7}
\multirow{2}{*}{}        & R@1   & R@5   & R@10  & R@1   & R@5   & R@10    \\
\midrule

w/o atk & 96.30 & 99.70 & 100.00 & 86.14 & 97.68 & 98.82 \\
0.5     & 33.50 & 51.10 & 60.50  & 27.98 & 50.72 & 59.06 \\
1       & 8.80  & 16.70 & 21.50  & 8.00  & 17.88 & 23.76 \\
2       & 1.10  & 1.80  & 3.40   & 1.72  & 4.14  & 6.08  \\
4       & 0.20  & 0.60  & 1.00   & 0.60  & 1.34  & 1.70  \\
8       & 0.10  & 0.20  & 0.20   & 0.14  & 0.52  & 0.82  \\
16      & 0.00  & 0.10  & 0.10   & 0.06  & 0.10  & 0.16 \\

\bottomrule
\end{tabular}
\end{table}

\begin{table}[t]
\caption{
The experimental results when \(step = 7\). The reported values are recall rates. Lower is better.
}
\label{tab:step7}
\begin{tabular}{c|cccccc}
\toprule
\multirow{2}{*}{\(\epsilon\ (\ell_\infty)\)}  &\multicolumn{3}{c}{Test Retrieval} & \multicolumn{3}{c}{Image Retrieval} \\
\cmidrule(r){2-4} \cmidrule(r){5-7}
\multirow{2}{*}{}        & R@1   & R@5   & R@10  & R@1   & R@5   & R@10    \\
\midrule

w/o atk & 96.30 & 99.70 & 100.00 & 86.14 & 97.68 & 98.82 \\
0.5     & 22.40 & 36.60 & 44.50  & 19.44 & 36.12 & 44.24 \\
1       & 1.80  & 4.50  & 6.10   & 2.62  & 5.88  & 8.58  \\
2       & 0.00  & 0.10  & 0.30   & 0.16  & 0.52  & 0.72  \\
4       & 0.00  & 0.00  & 0.00   & 0.00  & 0.00  & 0.02  \\
8       & 0.00  & 0.00  & 0.00   & 0.00  & 0.00  & 0.00 \\

\bottomrule
\end{tabular}
\end{table}

\begin{table}[b]
\caption{
The experimental results when \(step = 10\). The reported values are recall rates. Lower is better.
}
\label{tab:step10}
\begin{tabular}{c|cccccc}
\toprule
\multirow{2}{*}{\(\epsilon\ (\ell_\infty)\)}  &\multicolumn{3}{c}{Test Retrieval} & \multicolumn{3}{c}{Image Retrieval} \\
\cmidrule(r){2-4} \cmidrule(r){5-7}
\multirow{2}{*}{}        & R@1   & R@5   & R@10  & R@1   & R@5   & R@10    \\
\midrule

w/o atk & 96.30 & 99.70 & 100.00 & 86.14 & 97.68 & 98.82 \\
0.5     & 20.30 & 32.30 & 39.00  & 17.00 & 32.30 & 39.84 \\
1       & 1.00  & 3.40  & 4.60   & 1.96  & 4.62  & 5.96  \\
2       & 0.00  & 0.00  & 0.00   & 0.02  & 0.16  & 0.32  \\
4       & 0.00  & 0.00  & 0.00   & 0.00  & 0.00  & 0.00 \\

\bottomrule
\end{tabular}
\end{table}

\section{Combine FGA with unimodal attacks} \label{Combine FGA with unimodal attacks}
We design FGA as a universal attack strategy, which is theoretically orthogonal to all unimodal attack schemes.
\subsection{Global Perturbation} \label{Global Perturbation}
This subsection discusses how to generate global perturbations based on FGA.
We denote \(\delta\) as the added adversarial perturbation, with \(B(\epsilon, p) = \{\delta : \left\|\delta\right\|_p \le \epsilon\}\) representing the ball of perturbations bounded by \(\epsilon\) in \(p\)-norm.
Finding \(\delta\) typically can be addressed through an iterative process \cite{pgd,rq1,rq2,lwr3}, which can be summarized as three phases: obtaining gradient information (Eq~\ref{obtaining gradient information}), determining the steepest ascent direction (Eq~\ref{determining the steepest ascent direction}), and applying projection (Eq~\ref{applying projection}) \cite{fgsm}.

\textbf{(1) Obtaining gradient information:}
\begin{equation} \label{obtaining gradient information}
g = \nabla _{\delta^{(i)}}L_{gui}(v+\delta^{(i)}, W, Y)
\end{equation}
where \(\delta^{(i)}\) represents the perturbation at the \(i\)th iteration, \(W\) represents the guidance vectors, and \(Y\) represents the guidance labels. 
For simplicity, hereafter it is abbreviated as \(L_{gui}(v)\).

\textbf{(2) Determining the steepest ascent direction:}
\begin{equation} \label{determining the steepest ascent direction}
g^{(p)} = Dir_p(g)
\end{equation}
where $g$ represents the original gradient information, and $g^{(p)}$ is a unit vector under $\ell_p$ constraint, with $\left\|g^{(p)}\right\|_p=1$. So that $g^{(p)}$ represents the fastest loss rising direction under the $\ell_p$ norm constraint.
The steepest ascent directions for $\ell_1$ \cite{max}, $\ell_2$, and $\ell_\infty$ \cite{fgsm} are as follows: 

\begin{equation} \label{gete}
e_{i} = \left\{\begin{matrix}
  sign(g_{i})& \left | g_{i} \right | \ge \left | g \right |^{(q)}  & \\
  0 & \left | g_{i} \right | < \left  | g \right |^{(q)}  &
\end{matrix}\right.
\end{equation}

\begin{equation} \label{l1}
g^{(1)} = e / \left \| e \right \| _{1}
\end{equation}

\begin{equation} \label{l2}
g^{(2)} = g/\left \| g \right \| _2
\end{equation}

\begin{equation} \label{linf}
g^{(\infty)} = sign(g)
\end{equation}
where $\left  | g \right |^{(q)}$ denotes the $q^{th}$ percentile of $\left  | g \right |$.

\textbf{(3) Applying projection:}
\begin{equation} \label{applying projection}
\delta^{(i+1)}=Clamp_{(-v,1-v)}P_{B(\epsilon,p)}(\delta^{(i)}+\alpha\cdot g^{(p)})
\end{equation}
where \(P_{B(\epsilon,p)}\) ensures that \(\left \| \delta^{(i+1)} \right \|_p \le \epsilon\), and \(Clamp_{(-v,1-v)}\) ensures that the pixel values of \(v + \delta^{(i+1)}\) remain within the legal range \([0, 1]\).
To elaborate further, when moving $\delta^{(i)}$ along the steepest ascent direction with a step size of $\alpha$, it may lead to $\left\| \delta^{(i)}+\alpha\cdot g^{(p)} \right\|_p > \epsilon$. 
In such case, the projection algorithm is required to ensure $\left\| P_{B(\epsilon,p)}(\delta^{(i)}+\alpha\cdot g^{(p)} \right\|_p = \epsilon$. 
\begin{equation} 
P_{B(\epsilon,2)}(\delta)=\begin{cases}
  \epsilon\cdot\frac{\delta}{\left \| \delta \right \|_2 } & \text{ if } \left \| \delta \right \|_2>\epsilon \\
 \delta& \text{ if } \left \| \delta \right \|_2\le\epsilon
 \end{cases}
% P_{B(\epsilon,2)}(*)=* \cdot \frac{\epsilon }{Clamp_{(\epsilon,\infty )}(\left \| *\right \|_2)} 
\end{equation}

\begin{equation} 
P_{B(\epsilon,\infty)}(\delta)=Clamp_{(-\epsilon,\epsilon)}(\delta)
\end{equation}
where \(Clamp_{(-\epsilon,\epsilon)}\) represents clipping each element value in \(\delta\) to be between \(-\epsilon\) and \(\epsilon\). 
Besides, $P_{B(\epsilon,1)}$ involves a complex projection strategy for sparsity $\ell_1$ perturbation, discussed in detail in APGD\(_{\ell_1}\)\cite{apgd1} and MAX\cite{max}.

Based on what is mentioned above, we can execute global perturbation attacks FGA\(_{\ell_1}\), FGA\(_{\ell_2}\), FGA\(_{\ell_{\infty}}\) according to different norm constraints.
In the unimodal domain, multi-norm attacks are very necessary. 
This is because a classic defence strategy in the unimodal domain, adversarial training, often overfits adversarial examples of a certain norm. 
That is, it can effectively defend against adversarial examples of a specific norm but is ineffective against adversarial examples of other norms \cite{max}.
Therefore, the interplay of adversarial perturbations across multiple norms can better explore the lower bounds of a network's robustness \cite{msd, eat, ncat}.

\subsection{Momentum Mechanism} \label{Momentum Mechanism}
The momentum mechanism is a commonly used strategy to enhance the robustness of adversarial examples. 
On top of the global perturbation attack, it involves introducing momentum updates, during obtaining gradient information (Eq~\ref{obtaining gradient information}) \cite{mi}.

\textbf{Obtaining gradient information with momentum mechanism:}
\begin{equation}
g \gets \nabla _{\delta^{(i)}}L_{gui}(v+\delta^{(i)})
\end{equation}
\begin{equation}
g \gets g / mean(abs(g))
\end{equation}
\begin{equation} 
g \gets g + \alpha \cdot g_m
\end{equation}
\begin{equation} 
g_m \gets g
\end{equation}
where \(abs\) represents taking the absolute value of each element in \(g\), while \(mean\) denotes calculating the average of all element values.
\(g_m\) is initialized as an all-zero matrix, incorporating gradient information from previous iterations. Therefore, after introducing the momentum mechanism, the gradient information comes from the weighted sum of current gradient \(g\) and past gradient \(g_m\), with \(g_m\) weighting \(\alpha\).

\subsection{Patch Perturbation} \label{Patch Perturbation}
Global attacks constrain the perturbation \(\delta\) through \(\epsilon\), requiring the perturbation to be as small as possible to avoid human detection.
Patch attacks, on the other hand, use a binary mask matrix \(m\) to specify the patch's location information. Patch attacks concentrate the perturbation within a specified area of the image, typically a square, covering about 2\% of the original image's area \cite{lavan}.
Within this area, there's no need to limit the size of the perturbation, so no norm constraints are necessary. 
It's only required to ensure that the patch's pixel values are within the legal range \([0, 1]\).
Patch attack is also typically carried out in an iterative form:
\begin{equation}
g = \nabla _{\delta^{(i)}}L_{gui}(v \odot (1-m) + \delta^{(i)}\odot m)
\end{equation}
\begin{equation} 
\delta^{(i+1)} = Clamp_{(0,1)}(\delta^{(i)} + g)
\end{equation}
where \(\odot\) denotes the element-wise product, in the mask \(m\), an element value of 0 indicates that the original pixel at that position is replaced by a patch pixel.

\begin{figure}[t]
  \centering
  \includegraphics[width=\linewidth]{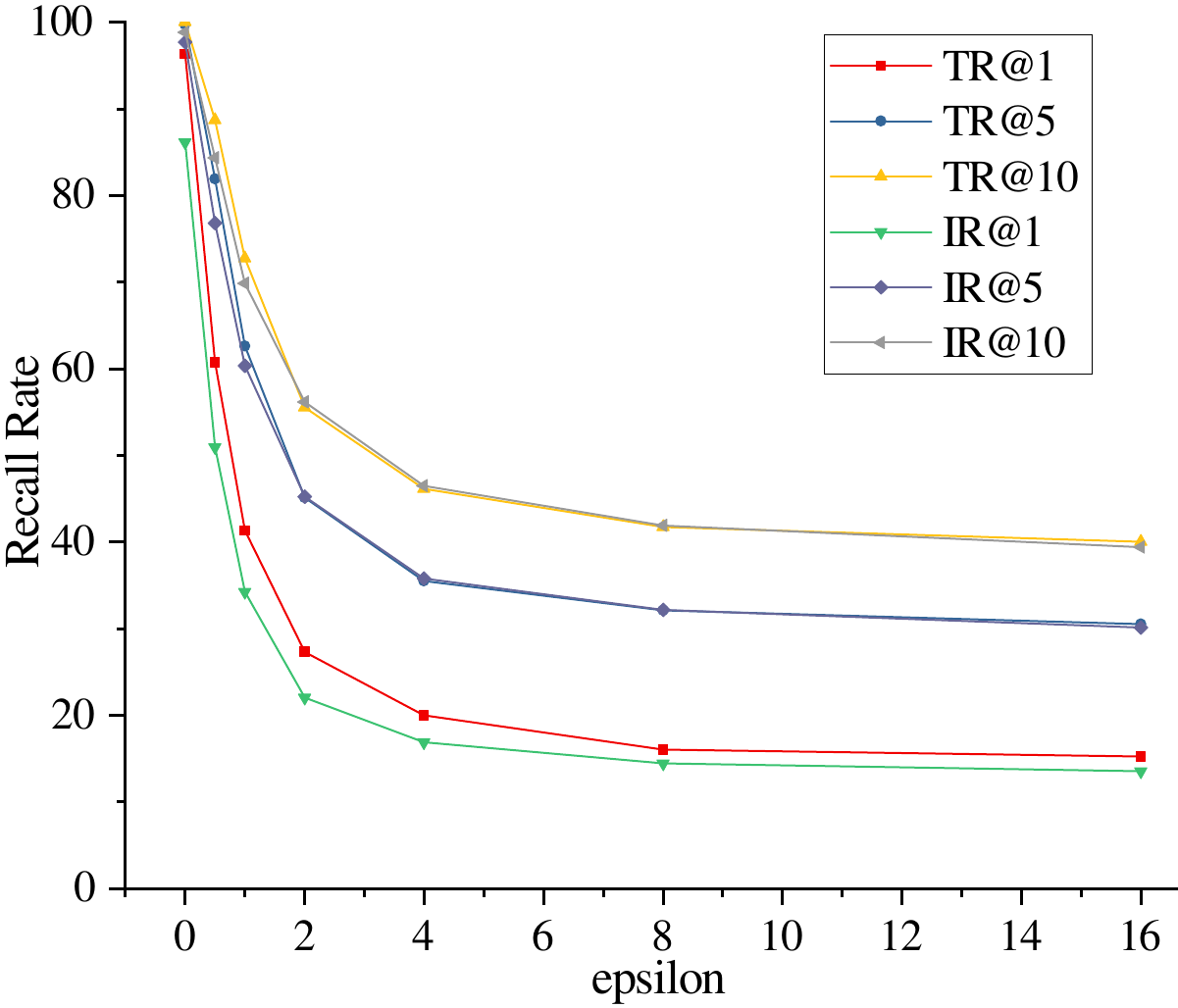}
  \caption{The experimental results when \(step = 1\). We observe that as the noise constraint is relaxed (with \(\epsilon\) increasing), the effectiveness of the attack gradually intensifies. However, the rate of decline in model performance slows down, indicating that the attack strength tends to converge.}
  \label{fig:step1}
% Adversarial image is achieved by guiding its embedding away from the aligned space into a non-aligned space, while ensuring the embedding does not approach the adversarial text embeddings.}
  % \vspace*{\fill}
  \Description{The experimental results when \(step = 1\).}
\end{figure}

\begin{figure}[t]
  \centering
  \includegraphics[width=\linewidth]{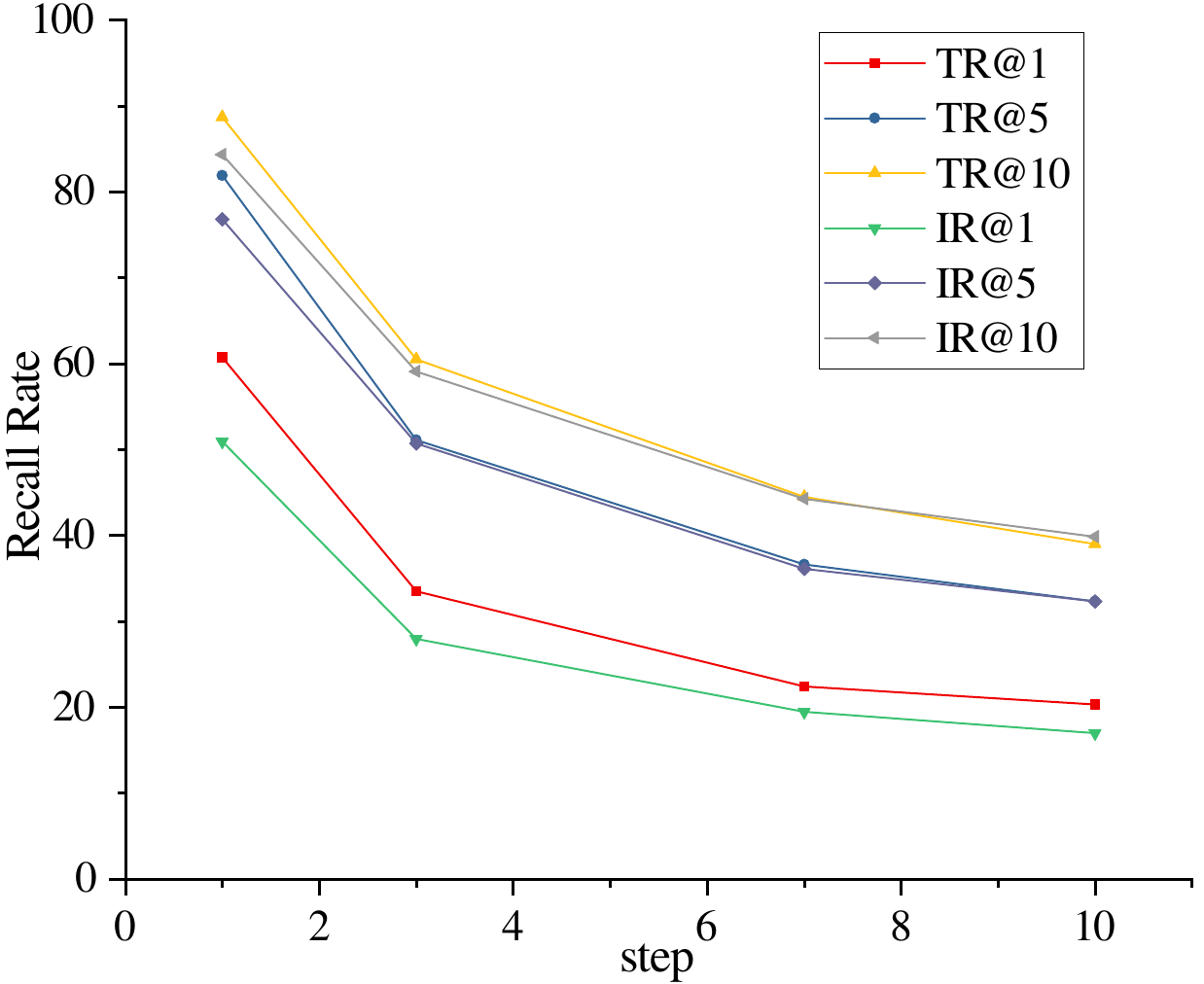}
  \caption{The experimental results when \(\epsilon = 0.5\). We note that as the number of iterations increases, the attack's effectiveness progressively intensifies. Nonetheless, the decrease in model performance decelerates, suggesting a convergence in attack potency.}
  \label{fig:epsilon0.5}
% Adversarial image is achieved by guiding its embedding away from the aligned space into a non-aligned space, while ensuring the embedding does not approach the adversarial text embeddings.}
  % \vspace*{\fill}
  \Description{The experimental results when \(step = 1\).}
\end{figure}

\section{More ablation experiments} \label{More ablation experiments}
We perform ablation studies focusing on the iteration count (\(step\)) and the intensity of noise (\(\epsilon\)) leveraging the BEiT3 model configured for the Image-Text Retrieval (ITR) task. 
The experimentation utilizes the BEiT3 model, which has been specifically fine-tuned utilizing the Flickr30k dataset, and evaluates its performance against the same dataset. 
Our methodology involves deploying the FGA while adhering to the \(\ell_\infty\) norm constraint, denoted as FGA\(_{\ell_\infty}\).
The experimental setup varies the \(step\) parameter across a set \(\{1, 3, 7, 10\}\), with corresponding outcomes detailed in Tab~\ref{tab:step1}, Tab~\ref{tab:step3}, Tab~\ref{tab:step7} and Tab~\ref{tab:step10}, respectively.
Concurrently, we explore a range of \(\epsilon\) values set at \{0.5, 1, 2, 4, 8, 16\}, ensuring that \(\left\|\delta\right\|_{\infty} \le \epsilon\), where the \(\epsilon\) values are pre-normalization, indicating that image pixel values span a \([0, 255]\) range.
We observe that: 
(1) As the noise constraint is relaxed (with \(\epsilon\) increasing), the effectiveness of the attack gradually intensifies (Fig~\ref{fig:step1}).
However, the rate of decline in model performance slows down, indicating that the attack strength tends to converge. (2) As the number of iterations (\(step\)) increases, the attack's effectiveness progressively intensifies (Fig~\ref{fig:epsilon0.5}).

%%
%% End of file `sample-sigconf-biblatex.tex'.

% \section{Attack Details} \label{sec:Attack Details}

% \subsection{Visual Question Answering}

% \subsection{Visual Grounding}

% \section{Combine FGA with unimodal attacks} \label{Combine FGA with unimodal attacks}

% \subsection{Momentum Mechanism} \label{Momentum Mechanism}

% \subsection{Patch Perturbation} \label{Patch Perturbation}

% \section{More ablation experiments} \label{More ablation experiments}

\end{document}